\documentclass[]{fairmeta}

\usepackage{xspace}
\usepackage{amsmath,amsfonts}
\usepackage{adjustbox}

\DeclareRobustCommand\onedot{\futurelet\@let@token\@onedot}
\def\@onedot{\ifx\@let@token.\else.\null\fi\xspace}
    \def\eg{\emph{e.g.},\,}

\usepackage{lineno}

\usepackage{enumitem}
\usepackage{todonotes}
\usepackage{algorithm}
\usepackage{algpseudocode}
\usepackage[title]{appendix}

\newcommand{\method}{\textsc{MT2IE}}
\newcommand{\methodlong}{Multimodal Text-to-Image Eval (\textsc{MT2IE})}

\definecolor{burntOrange}{RGB}{255,122,20}
\definecolor{forestGreen}{RGB}{33, 125, 12}
\definecolor{babyBlue}{RGB}{13, 138, 191}
\definecolor{darkRed}{RGB}{191, 25, 13}
\definecolor{addGreen}{RGB}{30,191,105}
\definecolor{antiqueone}{RGB}{133,92,117}
\definecolor{crimsonglory}{rgb}{0.75, 0.0, 0.2}
\definecolor{saffron}{RGB}{227, 170, 0}
\definecolor{brightpink}{RGB}{235, 52, 204}

\newcommand{\antiqueone}{\color{antiqueone}}

\newcommand{\removalRed}{\color{crimsonglory}}

\usepackage{tikz}
\usetikzlibrary{3d}
\usetikzlibrary{arrows.meta}
\usetikzlibrary{backgrounds}
\usetikzlibrary{bayesnet}
\usetikzlibrary{calc}
\usetikzlibrary{calligraphy}
\usetikzlibrary{fit}
\usetikzlibrary{positioning}
\usetikzlibrary{patterns}
\usetikzlibrary{decorations.markings}
\usetikzlibrary{decorations.pathreplacing}
\usetikzlibrary{shapes}
\usetikzlibrary{shapes.geometric}
\usetikzlibrary{shapes.multipart}
\usetikzlibrary{tikzmark}
\usetikzlibrary{shapes}
\usetikzlibrary{shapes.geometric}

\title{Multimodal Language Models as Text-to-Image Model Evaluators}

\author[1, 2, *]{Jiahui Chen}
\author[1]{Candace Ross}
\author[1]{Reyhane Askari-Hemmat}
\author[1]{Koustuv Sinha}
\author[1]{Melissa Hall}
\author[2]{Amy Zhang}
\author[1]{Michal Drozdzal}
\author[1, 3, 4, 5]{Adriana Romero-Soriano}

\affiliation[1]{FAIR at Meta}
\affiliation[2]{University of Texas at Austin}
\affiliation[3]{Mila}
\affiliation[4]{McGill University}
\affiliation[5]{Canada CIFAR AI chair}

\contribution[*]{Work done at Meta}

\abstract{The steady improvements of text-to-image (T2I) generative models lead to slow deprecation of automatic evaluation benchmarks that rely on static datasets, motivating researchers to seek alternative ways to evaluate T2I progress.
We present \textbf{\methodlong}, an evaluation framework in which a single multimodal large language model (MLLM) acts as an evaluator agent, iteratively generating the evaluation prompts and scoring the resulting images.
We show that \method's image-text consistency scores have higher correlation with human judgment than metrics previously introduced in the literature.
\method\ generates prompts that are efficient at probing T2I model performance: closely recovering the official T2I model rankings of three structurally distinct benchmarks from just 20 generated evaluation prompts, $28$--$105\times$ fewer than the benchmarks' own prompt sets.
When compared to existing evaluation metrics such as CLIPScore, VIEScore, and VQAScore, \method's T2I model rankings are more faithful and far more consistent across multiple evaluation seeds when using the same number of prompts.
\method\ can also \emph{adapt} evaluation to the model being tested: rewriting each prompt based on the model's own measured performance to produce a bespoke per-model benchmark that still recovers the official rankings and keeps the evaluated model in an informative scoring range.
We hope that these results will encourage the development of dynamic and interactive evaluation frameworks, and mitigate the deprecation of automatic evaluation benchmarks. \looseness-1
}

\date{August, 2026}
\correspondence{\email{ccross@meta.com}}

\begin{document}

\maketitle

\section{Introduction}\label{sec:intro}

With the ever-growing use of generated images in academic and commercial settings \citep{Rombach_2022_CVPR, podell2023sdxl, texttostickerstyletailoringlatent}, the evaluation of text-to-image (T2I) generative models is a rapidly growing and highly relevant area of research. This work focuses on evaluating image-text consistency: how well a generated image matches its corresponding text prompt.
Consistency directly impacts the overall quality of a generated image \citep{pareto, imagereward, pickscore}, and its assessment has gained increasing attention as users' text prompts become longer and more semantically complex.\looseness-1

Generally, \emph{static sets} of text prompts are used to benchmark T2I models.
These prompts are either drawn from the real world, typically sourced from image captioning datasets (\eg COCO~\citep{lin2014microsoft}, Conceptual Captions \citep{changpinyo2021conceptual}), or designed to test specific image generation capabilities (\eg
 ImagenHub~\citep{ku2023imagenhub}, ConceptMix \citep{wu2024conceptmix}, T2I-CompBench~\citep{t2icompbench}, HPDv2~\citep{human_pref_score}); in both cases, generated images are most commonly assessed with automatic metrics. 
For image-text consistency evaluation, \emph{alignment-based} or \emph{generated question-answering-based} (GQA) metrics are commonly used. \textit{Alignment-based} metrics judge image-text consistency by computing the similarity between the generated image and its prompt. For example, CLIPScore~\citep{clip_score} leverages the CLIP model~\citep{clip} to calculate cosine similarity between generated images and text prompts. However, its effectiveness is limited by the weaknesses of CLIP, which struggles with capturing compositional vision-language aspects such as spatial relationships, object cardinality, and attribute binding \citep{eyeswideshut, winoground, vlm_bagofwords}. \textit{GQA} metrics~\citep{tifa,dsg,vqascore} have been introduced to address these limitations and enable finer-grained image-text consistency evaluation. These use large language models (LLMs) to generate question sets about the text prompt and then use visual question answering (VQA) models to score the generated questions given the prompt's generated image.\looseness-1

\begin{figure}[t]
    \vspace{-0.8cm}
    \begin{center}
    \includegraphics[width=1.0\textwidth]{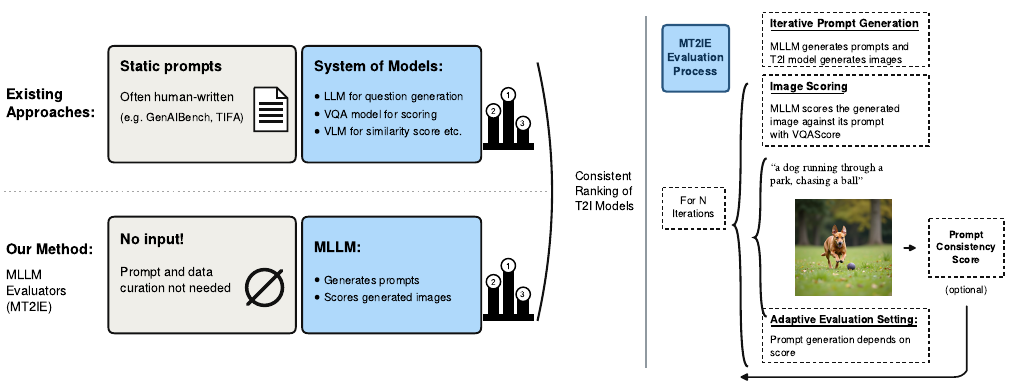}
    \end{center}
    \caption{We present \methodlong, our novel evaluation methodology for text-to-image (T2I) generation. Previous approaches to T2I model evaluation require large, static benchmarks and often multiple external models. Our approach is a single system that generates prompts and scores the resulting images. We show that \method\ can generate just 20 prompts while still closely recovering the official model rankings of approaches that use far more data.
    }
    \label{fig:method}
    \vspace{-0.3cm}
\end{figure}

Recently, multimodal LLMs (MLLMs) \citep{qwen3vl, llava, molmo} have been developed to enhance large language models (LLMs) with vision understanding capabilities, resulting in state-of-the-art performance across some vision-language tasks.
Various approaches use MLLMs for image-text consistency evaluation, often via some form of visual question answering~\citep{llmscore, viescore, gpt4_evaluator,meng2024image,tan2024evalalign}.
Although these approaches leverage MLLMs for improved evaluation, they still require fixed prompt datasets without any ability for dynamic evaluation.
This can create a narrow focus on optimizing T2I performance for predefined prompts, rather than driving broad improvements.
Additionally, rapid advancements in model capabilities can render static prompt sets ineffective, as they become overly simplistic and fail to provide meaningful evaluation~\citep{datasetdeprecation,ott2022mapping,gupta2024improving}.
Across benchmarks and metrics, the use of \textit{static evaluation frameworks} is prone to overfitting and vulnerable to deprecation.

In this work, we present \textbf{\methodlong}, a unified framework that moves beyond static evaluation by iteratively generating dynamic prompts and scoring image generations. 
We explore the potential of MLLMs as \emph{evaluator agents} that interact with a T2I model. Instead of relying on a static benchmark of prompts for image generation, the MLLM is now responsible for both \textit{generating} text prompts and \textit{scoring} the generated images. 
This removes the need to curate a prompt set for benchmarking while performing end-to-end evaluation with a single model.
This additionally allows us to leverage the MLLM to generate \textit{increasingly complex} prompts for generation, pushing the limits of the T2I model, and \textit{adaptive} prompts, automatically rewriting prompts according to the T2I model's performance.
We use only off-the-shelf open-source MLLMs, in contrast with previous work that either relied on proprietary GPT-4 variants \citep{gpt4_evaluator} or found open-source MLLMs' judgments ineffective and uncorrelated with human judgments \citep{viescore, genai_arena}.
Ranking 8 recent T2I models across three structurally distinct benchmarks (GenEval \citep{geneval}, T2I-CompBench \citep{t2icompbench}, GenAIBench \citep{genaibench}), we find that \method\ closely recovers each benchmark's official model ranking from a single 20-prompt run---more faithfully and far more consistently than existing metrics given the same prompt budget.
Our contributions are as follows:\looseness-1

\begin{enumerate}
    \item We validate MLLMs as image-text consistency judges: on the TIFA160 benchmark \citep{tifa}, a single open-source MLLM correlates with human judgment better than every automatic metric and multi-model GQA pipeline we compare against, replacing the separate question-generation, filtering, and VQA models those pipelines require.

    \item We show that the same MLLM can \emph{generate} the benchmark: \method\ closely recovers the T2I model rankings from three widely used benchmarks using only 20 generated evaluation prompts ($28$--$105\times$ fewer than the benchmarks' own prompt sets), and does so at roughly half the run-to-run standard deviation of existing metrics given the same prompt budget.

    \item We show that \method\ can \emph{adapt} evaluation to each T2I model by rewriting prompts based on that model's own measured performance, producing bespoke per-model evaluation trajectories that still recover the official benchmark rankings. Additionally, this adaptive approach keeps every T2I model in an informative scoring range as the prompt budget grows.
    To our knowledge, we are the first to adapt generated T2I evaluation prompts to the model at evaluation time in this way.
\end{enumerate}

We present these contributions as three experiments, each building on the last.
Section~\ref{sec:contribution_1} validates MLLMs as image-text consistency judges and fixes the scorer we use throughout.
\textbf{Section~\ref{sec:mllm_genbench} introduces the \method\ framework}, then tests
the prompts it generates against three widely used static benchmarks.
Then finally, Section~\ref{sec:adaptive} adds per-model adaptation. Appendix~\ref{app:aesthetic_evals} shows that \method\ transfers to a different evaluation axis, image aesthetics, by changing only the MLLM system prompt.\looseness-1

\section{Related Work}\label{sec:related_work}
\paragraph{Alignment-based image-text consistency metrics.} Alignment-based metrics use similarity scores between image and text in the embedding space of a vision-language model (VLM). VLMs such as CLIP \citep{clip} or BLIP \citep{blip} are commonly used as automatic image-text consistency metrics.
The most long-standing of these metrics is CLIPScore \citep{clip_score}, which computes the cosine similarity of the image and text CLIP embeddings.
However, many works have shown that CLIP and other VLMs treat text as a bag-of-words, failing to capture the semantics of compositional text that includes multiple objects and their attributes and spatial relations \citep{vlm_bagofwords, vlm_bad_composition, winoground, vlm_composition_bottleneck}.
Thus their similarity estimates are not reliable, especially when text prompts specify cardinality or compositional reasoning \citep{equibench, vlm_bad_composition, vlm_composition_bottleneck}.
Despite this, many recent, impactful image and video generation methods still use CLIPScore as an image-text consistency metric \citep{dreambooth, instruct_pix2pix, muse, sdedit, gans_t2i, imagen_video, tuneavideo}.
Similarity-based approaches have been enhanced by collecting large scale human ratings of generated images to enable finetuning of CLIP~\citep{pickscore, human_pref_score}. Similarly, human preference data has been used to create reward models out of alternative VLMs such as BLIP~\citep{imagereward}. Of these methods, it is unclear if any have achieved as widespread adoption as CLIPScore.\looseness-1

\paragraph{Large language models as evaluators.} Following the use of LLMs for image-text consistency question generation, LLMs or multimodal LLMs are also being used as standalone evaluators \citep{llmscore, viescore, gpt4_evaluator,tan2024evalalign}. \citet{llmscore} use a VLM to generate a text description of a generated image, then ask an LLM to judge the faithfulness of the image description and the image prompt. \citet{viescore} and \citet{gpt4_evaluator} prompt MLLMs for image-prompt faithfulness and overall image quality scores.
Both works focus on using proprietary, multimodal versions of GPT-4, finding that the MLLM used has great effect on correlation with human evaluators.
Notably, \citet{viescore} find that open-source MLLMs are significantly worse at generated image evaluation, with LLaVA \citep{llava} having 90\% worse human correlation compared to GPT-4v when using their evaluation methodology.\looseness-1

\paragraph{Dynamic and generated benchmarking.} A growing line of work moves beyond administering one large, human-curated test set to every model, either by \emph{automatically generating} evaluation content---in some cases evolving it with the models being measured---or by reducing an existing benchmark to a small, informative subset.
Most of this work targets language rather than T2I evaluation. Adversarial benchmarks have human annotators write examples that fool a target model, tying benchmark difficulty to model capability \citep{dynabench, anli, beattheai}. LLM-driven frameworks automate that loop, generating evaluation items and adapting their difficulty to the model under test, either by searching for high-error dataset descriptions \citep{autobencher} or by controlling the complexity of synthetically generated samples \citep{dyval}. A third line of this LLM evaluation work leaves the benchmark fixed but compresses it \citep{tinybenchmarks, anchorpoints, rodriguez2021evaluation}, though subset selection requires per-item results over the full benchmark first.
Within T2I evaluation, generated question answering (GQA) metrics use a language model to generate scoring questions for human-curated prompts and a VQA model to answer them using the generated image \citep{dsg, tifa, vq_squared_(baseline), vq_2, vis_prog}; while more fine-grained than alignment metrics, they are not necessarily more reliable, exhibiting biases from the models used to score questions \citep{saxon2024evaluates, ros2024smakes, goyal2017making}. Other T2I benchmarks generate the \emph{prompts} themselves, such as ConceptMix \citep{wu2024conceptmix}, which uses an LLM to compose prompts of controllable difficulty and scores each image with a \emph{separate} vision-language model answering one question per concept.
In each of these T2I approaches, generation and scoring are performed by separate models and the evaluation content is generated independently of the model being evaluated, so the same benchmark is administered to every T2I model. In contrast, \method\ uses a \emph{single} MLLM to both generate the evaluation prompts and score the resulting images, and its adaptive setting tailors prompt difficulty to \emph{each model} from that model's own measured performance, assuming neither a curated prompt pool nor prior results.





\section{Experiment 1: MLLMs for Image-Text Consistency Scoring}\label{sec:contribution_1}

A consistency benchmark scores how well each generated image matches its prompt; this scoring or judging of image-text consistency is the component we validate first.
We test whether off-the-shelf, open-source MLLMs can serve as such judges, reproducing the results of widely used scoring approaches on a fixed set of existing benchmark prompts. We use 5 different MLLMs which span a range of sizes, release dates, and training paradigms: Molmo-7B-D \citep{molmo}, Llama-3.2-11B-Vision-Instruct \citep{llama3.2_vision}, LLaVA-v1.6-34B \citep{llava}, Qwen3-VL-8B-Instruct, and Qwen3-VL-32B-Instruct \citep{qwen3vl}.
We selected only open-source models, as existing work on MLLMs as T2I evaluators either uses proprietary models \citep{viescore, gpt4_evaluator} or fine-tunes open-source ones for the task \citep{tan2024evalalign}, and we want to demonstrate that off-the-shelf open-source MLLMs work well in our framework without any fine-tuning.

We evaluate each MLLM's ability to score the TIFA160 benchmark \citep{tifa} using two different scoring approaches applied to the same fixed set of prompts: (i) a generated question answering (GQA) approach, where questions about a given prompt are generated and answered by the MLLM (\eg \citep{tifa,dsg, vq_2, vq_squared_(baseline), vp_eval}) and (ii) a VQAScore-based approach \citep{vqascore}, which computes the likelihood of answering ``Yes'' to the question \textit{``Does this figure show \{text prompt\}? Please answer yes or no.''}.
Each MLLM scores the TIFA160 prompts and their generated images with both approaches, and we then compare which scoring method and MLLM achieves the highest correlation with human judgments of image-text consistency.

\subsection{Experimental Setup}\label{sec:exp1_setup}

We use images and human ratings from prior work \citep{dsg} on the TIFA160 benchmark, which contains 160 prompts evaluated on 5 T2I models: LDM1.1 \citep{Rombach_2022_CVPR}, LDM1.5 \citep{Rombach_2022_CVPR}, LDM2.1 \citep{Rombach_2022_CVPR}, minDALL-E \citep{kakaobrain2021minDALL-E}, and VQ-Diffusion \citep{gu2022vector}, resulting in 800 image-text pairs with human consistency ratings on a 1--5 Likert scale. We measure the correlation of each scoring method with human ratings using Spearman's $\rho$ and Kendall's $\tau$.

For the \textbf{GQA approach}, each MLLM generates multiple-choice questions from the text prompt, filters them for validity, and then answers them given the generated image. The VQA accuracy serves as the consistency score. The system prompts used for question generation are in Appendix \ref{app:vqa_acc_with_prompts}.
We first verify that all MLLMs have strong VQA performance by evaluating them on TIFAv1.0's original, GPT-3-generated question set on real COCO images; results in Figure \ref{fig:coco_vqa} (Appendix \ref{app:vqa_acc_with_prompts}) confirm that all MLLMs outperform the VQA models used in existing GQA benchmarks.
We additionally compare the GPT-3-generated questions in TIFA160 and the MLLM-generated questions from LLaVA-v1.6-34B and find that entities in both question sets are very similar, validating that MLLM-generated questions are comparable to LLM-generated ones. Detailed results on generated question set verification are in Appendix \ref{app:llava-tifa}.

For the \textbf{VQAScore approach} \citep{vqascore}, we compute the probability of the MLLM generating ``\textit{yes}'' when asked \textit{``Does this figure show \{text prompt\}? Please answer yes or no.''} VQAScore has been proposed as an alternative to GQA-based approaches and alignment metrics like CLIPScore, motivated by CLIPScore's challenges with object composition~\citep{vqascore, eyeswideshut, winoground, vlm_bagofwords}, and the tendency of GQA-based approaches to exhibit a ``yes-bias'' in question answering tasks \citep{ros2024smakes}.\looseness-1

\subsection{Results}\label{sec:exp1_results}

\begin{table}[t]
\scriptsize
\setlength{\tabcolsep}{4pt}
\renewcommand{\arraystretch}{0.9}
\begin{center}
\caption{
Human judgment correlations on TIFA160 (Spearman's $\rho$, Kendall's $\tau$), for existing GQA (generated question answering) benchmarks (top), automatic metrics and VQAScore baselines (middle), and open-source MLLMs under both scoring approaches (bottom). Automatic metrics, like VQAScore, produce a single score per image-text pair. Every MLLM outperforms all existing GQA benchmarks and all training-free automatic metrics; Qwen3-VL-32B additionally outperforms the fine-tuned CLIP-FlanT5 VQAScore, the strongest baseline in the table.
}
\label{tab:human_correlation}
\vspace{-0.1cm}
\begin{tabular}{@{}lcccc@{}}
\toprule[1.2pt]
 & \multicolumn{2}{c}{\textbf{GQA}} & \multicolumn{2}{c}{\textbf{VQAScore}} \\
\cmidrule(lr){2-3} \cmidrule(lr){4-5}
 & $\rho$ & $\tau$ & $\rho$ & $\tau$ \\
\midrule
\multicolumn{5}{@{}l}{\textbf{Existing GQA Benchmarks}} \\
TIFA (Instruct-BLIP) & 46.0 & 36.0 & --- & ---\\
VQ\textsuperscript{2}A (Instruct-BLIP) & 42.6 & 35.2 & --- & ---\\
DSG (PaLI) & 57.1 & 45.8 & --- & ---\\
\midrule
\multicolumn{5}{@{}l}{\textbf{Automatic Metrics}} \\
BLEU-4 & \multicolumn{2}{c}{---} & 26.1 & 18.3\\
ROUGE-L & \multicolumn{2}{c}{---} & 34.2 & 23.9\\
METEOR & \multicolumn{2}{c}{---} & 37.9 & 26.9\\
SPICE & \multicolumn{2}{c}{---} & 32.9 & 23.6\\
CLIPScore (ViT-B-32) & \multicolumn{2}{c}{---} & 27.6 & 19.1 \\
VQAScore (CLIP-FlanT5, \textit{fine-tuned}) & \multicolumn{2}{c}{---} & 70.7 & 52.9 \\
\midrule
\multicolumn{5}{@{}l}{\textbf{Open-Source MLLMs (ours)}} \\
Llama-3.2-11B-Vision-Instruct & 57.9 & 46.3 & 60.6 & 43.7 \\
Molmo-7B-D & 58.7 & 45.9 & 54.3 & 38.4 \\
LLaVA-v1.6-34B & 60.7 & 47.6 & 73.5 & 54.4 \\
Qwen3-VL-8B-Instruct & 60.3 & 47.7 & 69.5 & 51.2 \\
Qwen3-VL-32B-Instruct & \textbf{65.8} & \textbf{52.3} & \textbf{74.6} & \textbf{56.3} \\
\bottomrule[1.2pt]
\end{tabular}
\end{center}
\vspace{-0.3cm}
\end{table}

Table \ref{tab:human_correlation} presents the full comparison. Several findings emerge:
\textbf{MLLMs can replace multi-model GQA pipelines.} All 5 MLLMs achieve higher GQA correlations than existing benchmarks that rely on separate LLM and VQA models, demonstrating that a single open-source MLLM can execute end-to-end question generation and scoring.
\textbf{VQAScore and GQA are complementary.} Neither scoring approach strictly dominates the other. VQAScore achieves higher $\rho$ for every model except Molmo, and for Kendall's $\tau$ it outperforms GQA on 3 of 5 models (LLaVA, Qwen3-VL-8B, and Qwen3-VL-32B). This interaction between scoring method and model likely arises because GQA performance depends on both the quality of generated questions and the model's VQA accuracy, whereas VQAScore relies only on the model's image-text consistency evaluation ability. Models that generate more discriminative questions (e.g., Molmo, the one model whose GQA correlations exceed its VQAScore correlations) benefit from the GQA pipeline, while VQAScore is better for models with strong image-text matching capabilities. \textbf{VQAScore is simpler and more practical for \method.} GQA requires question generation, filtering, and answer matching, each sensitive to prompt-engineering and answer-matching heuristics that vary across models. VQAScore needs only a single forward pass per image with a fixed prompt template, a decisive advantage for a framework like \method\ that scores many images across many rounds. \textbf{Qwen3-VL-32B achieves the best overall performance.} Qwen3-VL-32B attains the highest correlation of any MLLM across all metrics and scoring settings, and the next-best model is LLaVA-v1.6-34B. Unsurprisingly, our results suggest that larger MLLMs are stronger image-text consistency scorers.

We therefore adopt VQAScore for \method, evaluating it with Qwen3-VL-8B---the best-correlating model under 12B parameters---and Qwen3-VL-32B, two scales from one family to show the effect of model size.

\section{Experiment 2: MLLMs for Benchmark Generation via \method} \label{sec:mllm_genbench}

\vspace{-0.3cm}
\begin{figure}[ht]
    \begin{center}
    \vspace{-0.3cm}
    \begin{tikzpicture}[
        every node/.style={font=\footnotesize, inner xsep=3pt, inner ysep=1.5pt},
        node distance=0.7em and 0em  
        ]
        \node(iter0)[draw,dashed,rounded corners]{
            \parbox{46em}{\raggedright \textit{iteration 1:} a dog running through a park, chasing a ball \\ \textbf{Consistency Score: 0.977}
            \\[2ex]}
        };
        \node[right=of iter0] (img0) {  
            \includegraphics[width=4.5em]{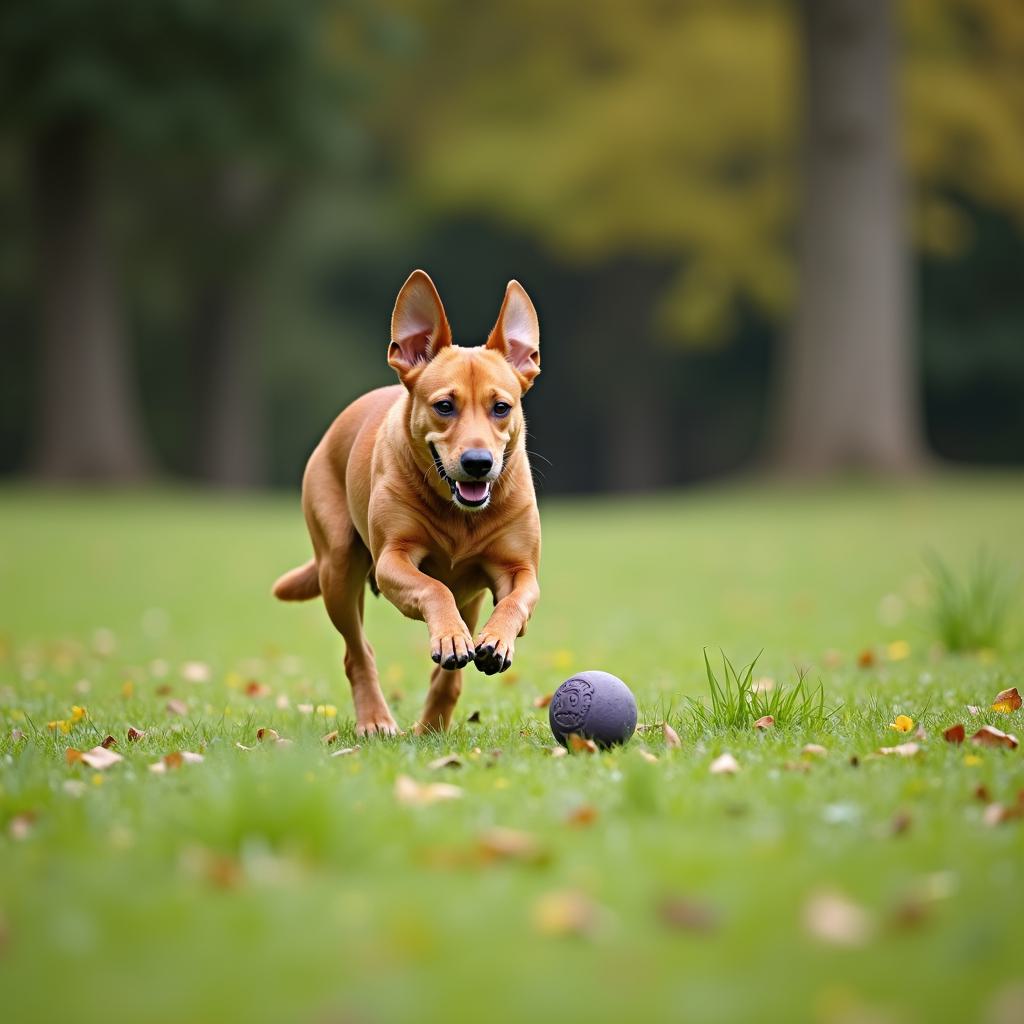}
        };

        \node(iter1)[draw,dashed,rounded corners,below=of iter0]{
            \parbox{46em}{\raggedright \textit{iteration 2:} a dog running through a park chasing a ball \textbf{\antiqueone with a frisbee in the air}\\ \textbf{Consistency Score: 0.938}
            } 
        };
        \node[right=of iter1] (img1) {
            \includegraphics[width=4.5em]{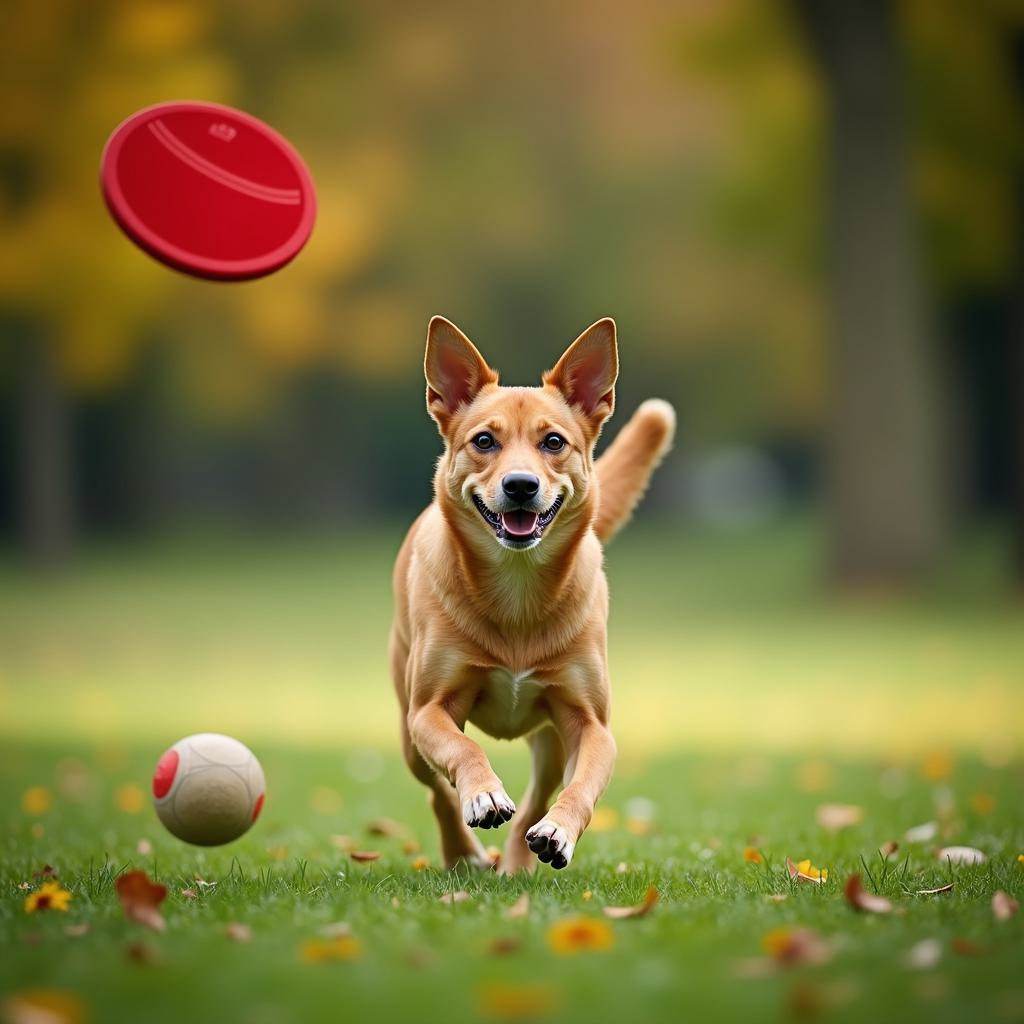}
        };

        \node(iter2)[draw,dashed,rounded corners,below=of iter1]{
            \parbox{46em}{\raggedright\textit{iteration 3:} a dog running through a park chasing a ball with a frisbee in the air \textbf{\antiqueone with a bird flying overhead}\\ \textbf{Consistency Score: 0.861}
            } 
        };
        \node[right=of iter2] (img2) {
            \includegraphics[width=4.5em]{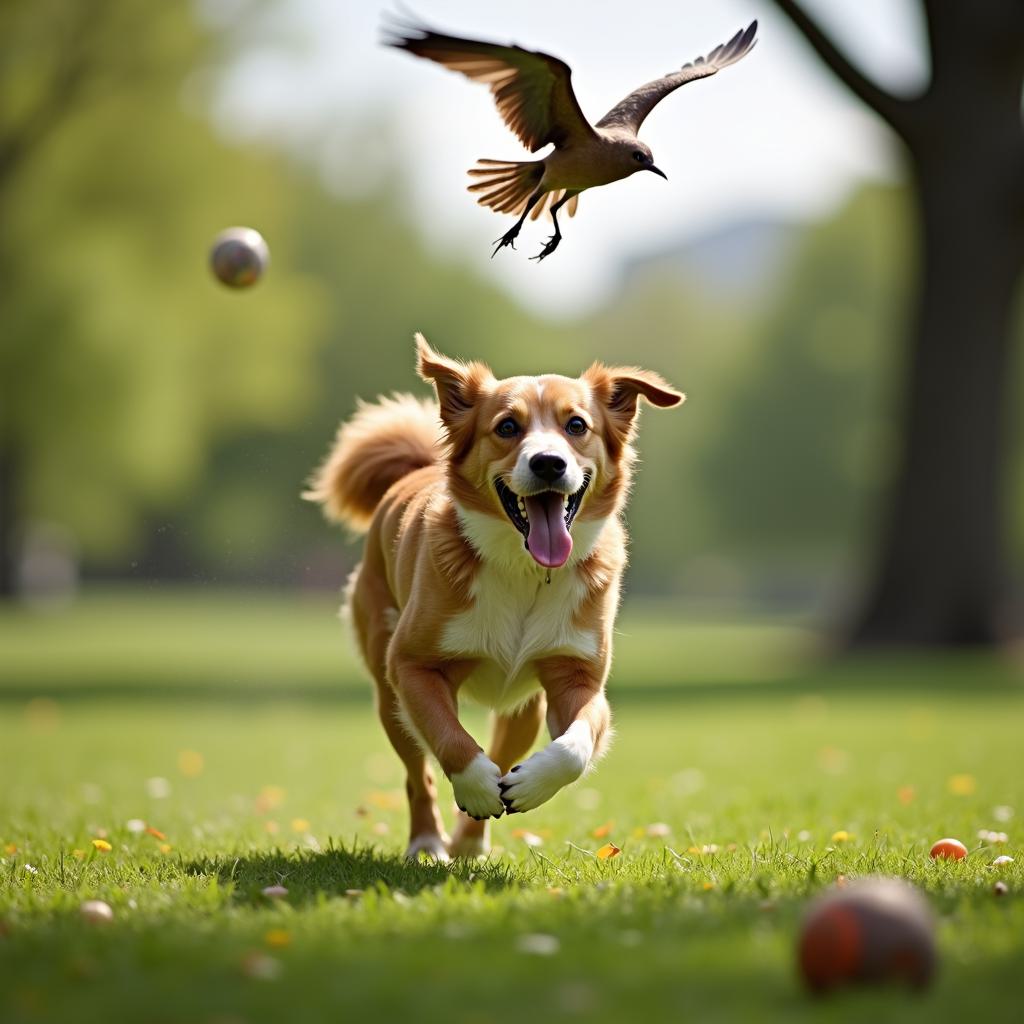}
        };

        \node(iter3)[draw,dashed,rounded corners,below=of iter2]{
            \parbox{46em}{\raggedright\textit{iteration 4:} a dog running through a park chasing a ball with a frisbee in the air with a bird flying overhead \textbf{\antiqueone and a person walking on a path in the background}\\ \textbf{Consistency Score: 0.944}
            }
        };
        \node[right=of iter3] (img3) {
            \includegraphics[width=4.5em]{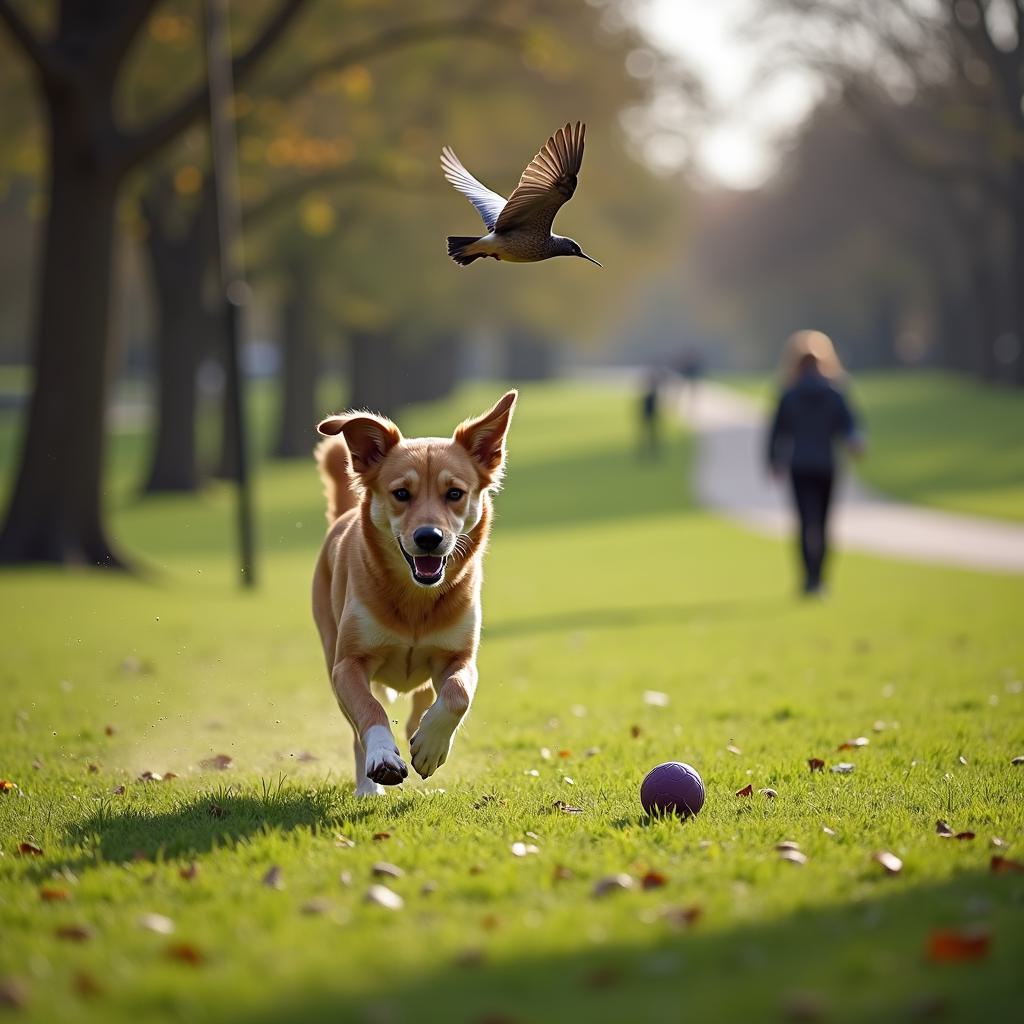}
        };

        \node(iter4)[draw,dashed,rounded corners,below=of iter3]{
            \parbox{46em}{\raggedright\textit{iteration 5:} a dog running through a park chasing a ball with a frisbee in the air a bird flying overhead and a person walking on a path in the background \textbf{\antiqueone with a small wooden bench on the side of the path}\\ \textbf{Consistency Score: 0.929}
            } 
        };
        \node[right=of iter4] (img4) {
            \includegraphics[width=4.5em]{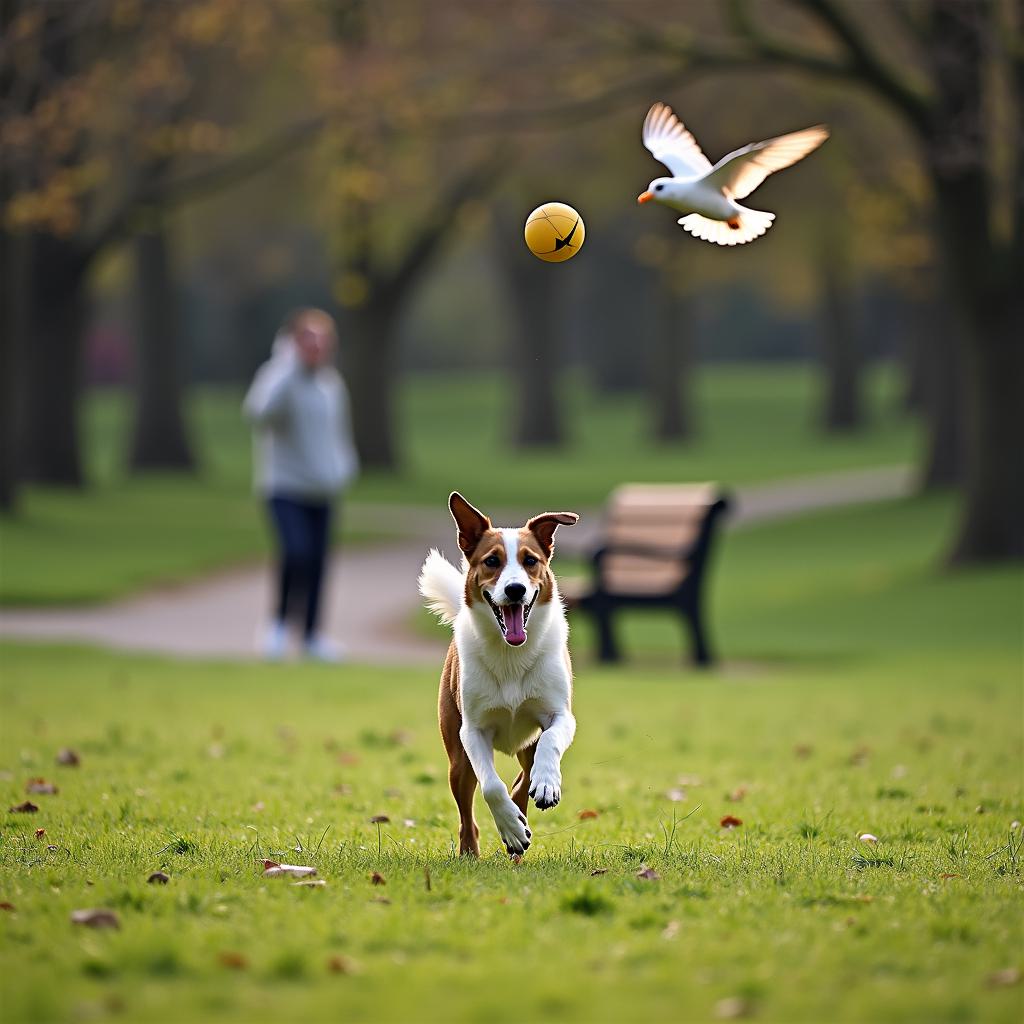}
        };

        \draw[->,line width=0.05mm] (iter0) -- (iter1);
        \draw[->,line width=0.05mm] (iter1) -- (iter2);
        \draw[->,line width=0.05mm] (iter2) -- (iter3);
        \draw[->,line width=0.05mm] (iter3) -- (iter4);
    \end{tikzpicture}
    \end{center}
    \caption{An example of our method \method\ generating progressively difficult prompts, with additions \textbf{\antiqueone{shown in purple}}.  Corresponding generated images and scores are also shown. Prompts become more complex while maintaining linguistic structure.}
    \label{fig:ex-iterative-prompts}
    \vspace{-0.4cm}
\end{figure}


Having validated in Section~\ref{sec:contribution_1} that MLLMs reliably \emph{score} image-text consistency on a given set of benchmark prompts, we now use the MLLM to \emph{generate} the prompts as well, so that a single model constructs the entire benchmark end-to-end.
We introduce our approach, \textbf{the \method\ framework}, and show that it can be used to iteratively generate efficient benchmarks for image-text consistency.
With each iteration, the generated prompts increase in complexity and thus challenge the T2I models in progressively more difficult settings.
\method-generated benchmarks are \textit{effective} in that they match prior model rankings from benchmarks of handwritten prompts, while also being \textit{efficient} in that they require far fewer examples than existing benchmarks.
Additionally, because \method\ generates its prompts rather than drawing from a fixed set, the benchmark can be regenerated to mitigate the dataset leakage and saturation that erode static benchmarks over time---a direction we develop with adaptive, per-model difficulty in Section~\ref{sec:adaptive}.

Throughout this section and Section~\ref{sec:adaptive} we evaluate the same 8 recent T2I models, spanning several architectures: the latent diffusion models (LDMs) SD2.1~\citep{Rombach_2022_CVPR}, SD3~\citep{sd3}, SDXL~\citep{podell2023sdxl}, and SDXL-Turbo~\citep{sauer2025adversarial}; the flow-based model Flux~\citep{flux}; the cascaded pixel-based diffusion model DeepFloyd~\citep{deepfloyd}; Playground 2.5~\citep{playground2.5}; and PixArt-$\alpha$~\citep{pixart-alpha}.
\vspace{-0.2cm}


\subsection{Progressive Difficulty Prompt Generation}\label{sec:framework}
Our approach iterates between two stages, shown on the right side of Figure \ref{fig:method}. First, during \textit{iterative prompt generation}, the MLLM is given an existing prompt $T_{i-1}$ and generates a new prompt $T_{i}$ that builds on $T_{i-1}$ to become more difficult.
$T_{i}$ contains an additional object, spatial relationship, or object attribute compared to $T_{i-1}$, incrementally increasing the complexity of the prompt every iteration; see Figure \ref{fig:ex-iterative-prompts} for an example.
Each run starts from a moderately complex (2--3 compositional concepts) seed prompt $T_{1}$, which may be user-written or MLLM-generated.
We generate one prompt for each of the 4 categories in COCO's captions~\citep{lin2014microsoft}: household objects and food, people, animals, and locations.
The T2I model generates an image $I_{i}$ based on the prompt $T_{i}$.
Next, during \textit{image scoring}, the MLLM scores $I_{i}$ for faithfulness to prompt $T_{i}$ with VQAScore \citep{vqascore} (Section~\ref{sec:exp1_results}).
A full run consists of $N$ iterations: the first scores the seed prompt $T_{1}$ directly, and each subsequent iteration $i$ rewrites $T_{i-1}$ into a harder $T_{i}$ before scoring it. A run therefore yields $N$ prompts $T_{1},\dots,T_{N}$ and $N$ scores $s_{1},\dots,s_{N}$, matching the iteration numbering in Figure~\ref{fig:ex-iterative-prompts}.
Because a single fixed schedule of increasing difficulty is applied to every T2I model, we refer to this as the \textbf{progressive difficulty} setting of \method, in contrast to the per-model adaptive setting we introduce in Section~\ref{sec:adaptive}.
A single \method\ benchmark spends its 20-prompt budget as 4 seed prompts---one per category---each escalated for $N=5$ iterations, favoring depth over breadth; $N=5$ is the best-performing setting for our headline Qwen3-VL-32B scorer in the iteration ablation of Section~\ref{sec:iter_ablation_body}. We note that this choice is made on the same rank-correlation signal we go on to report; Table~\ref{tab:iter_ablation} shows a broad plateau rather than a sharp optimum, so what matters is operating in the ${\sim}20$-prompt regime rather than the specific value of $N$.

To confirm that the loop produces progressively harder prompts, we measure how scores and prompt properties evolve across iterations. Averaged over all 8 T2I models and 5 MLLM seeds (100 generated prompts, 800 prompt--model scores), mean VQAScore falls monotonically across iterations, from $0.40$ at the first iteration to $0.04$ by the fifth, while mean prompt length grows from $16.4$ to $47.0$ words, showing that each rewrite reliably adds compositional content that T2I models find harder to render faithfully.
As \method-generated prompts get more difficult, they still maintain coherence and grammaticality (Figure~\ref{fig:ex-iterative-prompts}); we analyze prompt-difficulty metrics further in Appendix~\ref{app:prompt_complexity}.

\subsection{Efficient and Reliable Evaluation via \method}\label{sec:efficiency}

With the judge fixed to VQAScore, we now test the prompt component: do the prompts \method\ generates probe T2I models as effectively as the large, hand-curated prompt sets of static benchmarks?
We compare \method's ranking of 8 T2I models against the rankings from three structurally distinct and widely used benchmarks: GenEval~\citep{geneval} (553 prompts, object-detection accuracy via Mask2Former), T2I-CompBench~\citep{t2icompbench} (2,100 prompts, per-category BLIP-VQA / UniDet / CLIP metrics), and GenAIBench~\citep{genaibench} (1,600 prompts, CLIP-FlanT5 computed VQAScore).
Our ground truth is each benchmark's \emph{official ranking} of these 8 T2I models, recomputed by running the benchmark's released evaluation pipeline over its full prompt set since most models postdate the leaderboards. We rank the models with \method's generated prompts and measure Kendall's $\tau$ against it; this ground truth is external to every method, so \method\ and all baselines are held to the same standard.

\begin{table}[t]
\vspace{-0.8cm}
\footnotesize
\centering
\setlength{\tabcolsep}{8pt}
\caption{
\textbf{Recovering official benchmark rankings from 20 prompts.} Per-draw rank correlation (Kendall's $\tau$, mean\,$\pm$\,std over draws) between the T2I ranking each method produces from 20 prompts and each benchmark's official ranking: GenEval (native Mask2Former detection), T2I-CompBench (official per-category metric), and GenAIBench (CLIP-FlanT5 leaderboard). Baselines randomly sample their 20 prompts (10 seeds); \method\ generates them (5 seeds). \method's \emph{progressive} setting (this section) generates one prompt set per MLLM; its \emph{adaptive} setting (Section~\ref{sec:adaptive}) generates a bespoke set per T2I model and ranks models by difficulty-weighted consistency (each score weighted by prompt length). Higher mean and lower std are both better; \textbf{bold} marks the best mean per benchmark. Full per-count results are in Appendix~\ref{app:iter_ablation}.
}
\begin{tabular}{@{}lccc@{}}
\toprule[1.0pt]
\small
& \multicolumn{3}{c}{\textbf{Per-draw $\tau$ vs.\ Official Benchmark Ranking}} \\
\cmidrule(lr){2-4}
& GenEval & T2I-CompBench & GenAIBench \\
\midrule
\multicolumn{4}{@{}l}{\textit{Baselines (20 randomly sampled prompts):}} \\
\hspace{0.5em}CLIPScore (CLIP ViT-B/32) & $0.079${\scriptsize$\,\pm\,0.20$} & $0.293${\scriptsize$\,\pm\,0.19$} & $0.036${\scriptsize$\,\pm\,0.26$} \\
\hspace{0.5em}VIEScore (LLaVA-v1.6-13B) & $0.363${\scriptsize$\,\pm\,0.19$} & $0.587${\scriptsize$\,\pm\,0.10$} & $0.284${\scriptsize$\,\pm\,0.30$} \\
\hspace{0.5em}VQAScore (CLIP-FlanT5-XXL)$^{\dagger}$ & $0.514${\scriptsize$\,\pm\,0.21$} & $0.650${\scriptsize$\,\pm\,0.11$} & $0.443${\scriptsize$\,\pm\,0.20$} \\
\midrule
\multicolumn{4}{@{}l}{\textit{\method\ progressive difficulty (this section):}} \\
\hspace{0.5em}\method\ (Qwen3-VL-8B) & $0.614${\scriptsize$\,\pm\,0.12$} & $0.571${\scriptsize$\,\pm\,0.14$} & $0.571${\scriptsize$\,\pm\,0.08$} \\
\hspace{0.5em}\method\ (Qwen3-VL-32B) & $\mathbf{0.629}${\scriptsize$\,\pm\,0.11$} & $\mathbf{0.671}${\scriptsize$\,\pm\,0.06$} & $0.586${\scriptsize$\,\pm\,0.15$} \\
\midrule
\multicolumn{4}{@{}l}{\textit{\method\ adaptive, difficulty-weighted consistency (Section~\ref{sec:adaptive}):}} \\
\hspace{0.5em}\method\ (Qwen3-VL-8B) & $0.529${\scriptsize$\,\pm\,0.20$} & $0.629${\scriptsize$\,\pm\,0.16$} & $0.514${\scriptsize$\,\pm\,0.16$} \\
\hspace{0.5em}\method\ (Qwen3-VL-32B) & $0.557${\scriptsize$\,\pm\,0.24$} & $0.571${\scriptsize$\,\pm\,0.16$} & $\mathbf{0.600}${\scriptsize$\,\pm\,0.21$} \\
\bottomrule[1.0pt]
\end{tabular}
\\[2pt]
{\scriptsize $^{\dagger}$VQAScore shares its CLIP-FlanT5 evaluator with the GenAIBench leaderboard ground truth, giving it a built-in advantage on that column; both \method\ settings nonetheless exceed it there.}
\label{tab:ranking_corr}
\end{table}

\begin{table}[t]
\footnotesize
\centering
\setlength{\tabcolsep}{4pt}
\caption{
\textbf{Effect of the generated prompt count on \method.} As the number of difficulty iterations grows, we report $\bar{\tau}$ (per-draw Kendall's $\tau$ averaged over the three benchmarks' official rankings), the score spread (max$-$min mean VQAScore across the 8 T2I models), and the floor fraction (share of model--prompt scores $\le\!0.1$); see Section~\ref{sec:iter_ablation_body}. Qwen3-VL-32B peaks at 20 prompts and then saturates ($\bar{\tau}$ unchanged through 48); Qwen3-VL-8B is noisier and lower throughout. \textbf{Bold} marks each scorer's best $\bar{\tau}$. The full 11-iteration table is in Appendix~\ref{app:iter_ablation}.
}
\begin{tabular}{@{}cc ccc c ccc@{}}
\toprule[1.0pt]
& & \multicolumn{3}{c}{\textbf{Qwen3-VL-8B}} & & \multicolumn{3}{c}{\textbf{Qwen3-VL-32B}} \\
\cmidrule(lr){3-5} \cmidrule(lr){7-9}
\textbf{Iters} & \textbf{Prompts} & $\bar{\tau}$ & Spread & Floor & & $\bar{\tau}$ & Spread & Floor \\
\midrule
2  & 8  & $0.548$ & $0.25$ & $41\%$ & & $0.586$ & $0.36$ & $49\%$ \\
3  & 12 & $0.505$ & $0.26$ & $46\%$ & & $0.595$ & $0.35$ & $59\%$ \\
4  & 16 & $0.571$ & $0.24$ & $51\%$ & & $0.624$ & $0.31$ & $66\%$ \\
5  & 20 & $0.586$ & $0.22$ & $57\%$ & & $\mathbf{0.629}$ & $0.29$ & $70\%$ \\
6  & 24 & $0.581$ & $0.21$ & $62\%$ & & $0.629$ & $0.26$ & $74\%$ \\
8  & 32 & $\mathbf{0.600}$ & $0.17$ & $69\%$ & & $0.629$ & $0.21$ & $79\%$ \\
12 & 48 & $0.600$ & $0.13$ & $78\%$ & & $0.629$ & $0.15$ & $86\%$ \\
\bottomrule[1.0pt]
\end{tabular}
\label{tab:iter_ablation}
\vspace{-0.5cm}
\end{table}

\begin{figure}[t]
    \centering
    \includegraphics[width=\linewidth, clip]{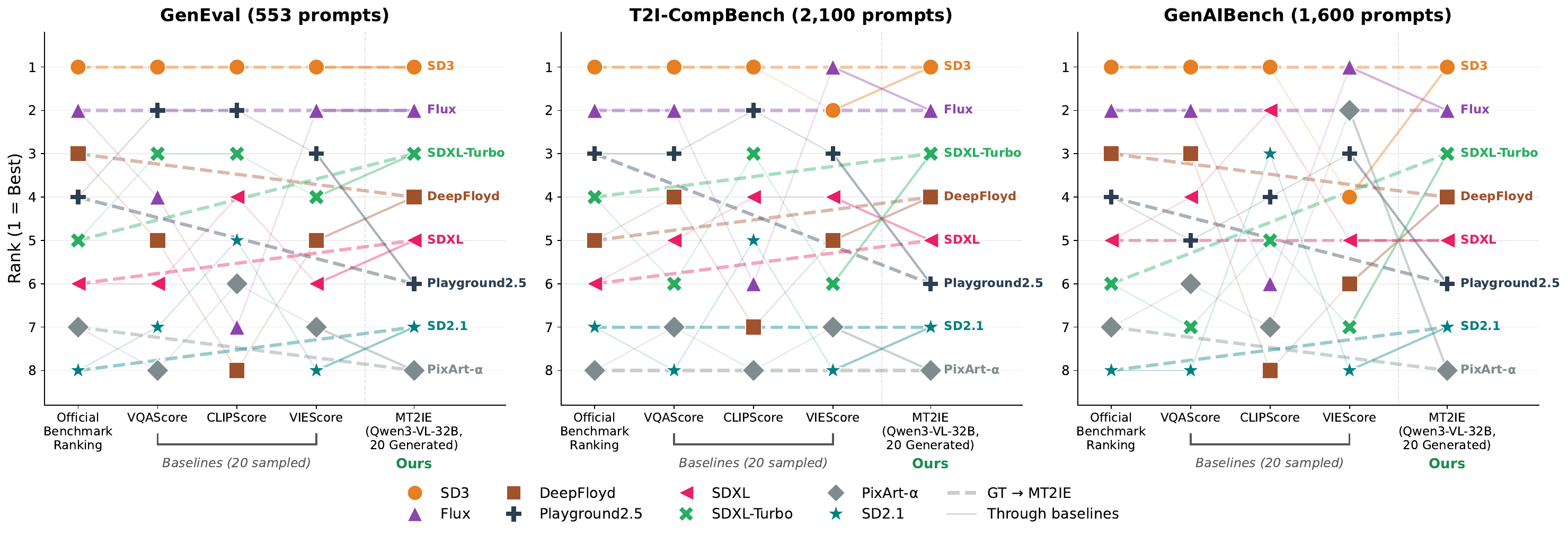}
    \caption{T2I model rankings, one panel per benchmark (rank 1=best, 8=worst). Within each panel, the \textit{leftmost column} is that benchmark's official ranking over its full prompt set; the \textit{three middle columns} are rankings from 20 randomly sampled prompts scored with VQAScore (CLIP-FlanT5), CLIPScore, and VIEScore; and the \textit{rightmost column} is \method's ranking from 20 prompts generated and scored by Qwen3-VL-32B. \textbf{Dashed lines} connect each model's official rank to its \method\ rank (near-parallel lines indicate preserved ordering) while solid lines trace ranks through the baselines, whose frequent crossings indicate scrambled orderings. Every ranking here pools scores over all draws (10 seeds per baseline, 5 seeds for \method); the per-draw correlations are reported in Table~\ref{tab:ranking_corr}.}
    \label{fig:all_rankings_genbench}
    \vspace{-0.4cm}
\end{figure}

We run \method\ with two MLLMs of different scales, Qwen3-VL-8B and Qwen3-VL-32B~\citep{qwen3vl}, and rank the 8 T2I models by each MLLM's average VQAScore on the generated prompts. As baselines, we randomly sample 20 prompts from each benchmark (uniformly across prompt categories when provided, otherwise uniformly at random) and rank the T2I models on them with three existing scorers, each using the model from its original work: VQAScore (CLIP-FlanT5-XXL), CLIPScore (CLIP ViT-B/32), and VIEScore (LLaVA-v1.6-13B).
Crucially, we report all correlations \emph{per-draw}: we correlate each individual prompt draw's ranking against the official ranking and report the mean and standard deviation across draws, rather than correlating a single ranking averaged over many draws. This measures what a practitioner experiences from one evaluation run and exposes the baselines' large run-to-run variance; because that variance is extremely high for the downsampled baselines, they are reported over 10 seeds and \method\ over 5. Results are shown in the first two blocks of Table~\ref{tab:ranking_corr}; its lower blocks report the adaptive setting introduced in Section~\ref{sec:adaptive}.

With Qwen3-VL-32B, \method\ recovers the official benchmark ranking more reliably than every baseline metric computed on the same number of prompts. Against VQAScore, the strongest baseline on all three benchmarks, it reaches Kendall $\tau$ of $0.629$ vs.\ $0.514$ on GenEval, $0.671$ vs.\ $0.650$ on T2I-CompBench, and $0.586$ vs.\ $0.443$ on GenAIBench. Qwen3-VL-8B still beats every baseline on GenEval and GenAIBench.
At $\tau\!\approx\!0.6$--$0.7$ this is recovery rather than exact reproduction of the official benchmark rankings.

\method\ is also markedly more \emph{consistent}. The baselines are not merely less accurate but wildly unstable: CLIPScore's std matches or exceeds its mean on GenEval and GenAIBench, making a single downsampled run there close to a coin-flip. Progressive \method\ runs at about half the baselines' average standard deviation ($0.11$ vs.\ $0.20$), so a \emph{single} 20-prompt run reliably lands near the official ranking---including on GenAIBench, where VQAScore has a built-in advantage (Table~\ref{tab:ranking_corr}, footnote).
Figure \ref{fig:all_rankings_genbench} shows the orderings themselves: \method's per-model ranks stay close to the official ranks (near-parallel dashed lines), whereas the baselines' ranks cross frequently, indicating scrambled orderings.
Notably, the 32B scorer's advantage holds across all three benchmarks despite their very different prompt structures (object-focused composition in GenEval, attribute binding and spatial relations in T2I-CompBench, and open-ended human-written prompts in GenAIBench), indicating that \method's generated prompts probe T2I performance across the axes these benchmarks span rather than overfitting to any one.
\method\ thus recovers a full benchmark's official model ranking from just 20 generated prompts---roughly $28\times$ fewer than GenEval's 553, $80\times$ fewer than GenAIBench's 1,600, and over $100\times$ fewer than T2I-CompBench's 2,100---making it far more efficient than the static benchmarks it recovers, and more reliable than downsampling them to the same budget.

\subsection{Difficulty Iterations Ablation}\label{sec:iter_ablation_body}
Table~\ref{tab:iter_ablation} ablates the number of difficulty iterations and therefore number of prompts used with \method. To summarize ranking quality with a single number per setting, we report $\bar{\tau}$, the per-draw Kendall's $\tau$ averaged over the three benchmarks' official rankings. Alongside it we track two properties of the underlying score distribution that do not depend on the ground-truth ranking: the \emph{score spread}, the gap between the highest- and lowest-scoring T2I model (max$-$min mean VQAScore), which measures how separable the models are; and the \emph{floor fraction}, the share of model--prompt scores at or below $0.1$, which measures how many generations the prompts drive to near-total failure.

Reading across iterations, the larger scorer, Qwen3-VL-32B, improves up to 20 prompts and then \emph{saturates}, with $\bar{\tau}$ unchanged from 20 through 48; the smaller Qwen3-VL-8B climbs more noisily, never reaches it, and gains only $0.014$ at its own best point of 32 prompts. Beyond ${\sim}20$ prompts, then, further generations cost compute without improving the evaluation. The spread and floor columns explain why: as the loop drives prompts harder, the score spread collapses ($0.36\!\rightarrow\!0.15$ for the 32B, $0.25\!\rightarrow\!0.13$ for the 8B) and the floor fraction climbs to $86\%$ and $78\%$, so the ranking comes to rest on a shrinking minority of comparisons between generations that have not already failed. The ordering itself survives over the range we test---$\bar{\tau}$ plateaus rather than falling---but the added prompts stop buying information: they push already-failing models further below the floor instead of discriminating among the strongest ones.
The deeper problem is that each T2I model has its own informative difficulty range, yet the progressive setting applies one schedule to all: by the time prompts are hard enough to discriminate among the strongest models, the weakest have long since saturated at their floor. This motivates tailoring difficulty per model, which we pursue with adaptive prompting in Section~\ref{sec:adaptive}.

\vspace{-0.2cm}
\section{Experiment 3: MLLMs as Adaptive Evaluator Agents}\label{sec:adaptive}
\vspace{-0.3cm}

\begin{figure}[ht]
    \vspace{-0.2cm}
    \centering
    \begin{tikzpicture}[
        every node/.style={font=\footnotesize, inner xsep=3pt, inner ysep=1.5pt},
        node distance=0.7em and 0em]

        \node(iter0)[draw,dashed,rounded corners]{
            \parbox{46em}{\raggedright \textit{iteration 1:} a set of knives and forks leaning against each other on a table \\\textbf{Score: 0.912} \\ [1.5ex]}
        };
        \node[right=of iter0] (img0) {
            \includegraphics[width=4.5em]{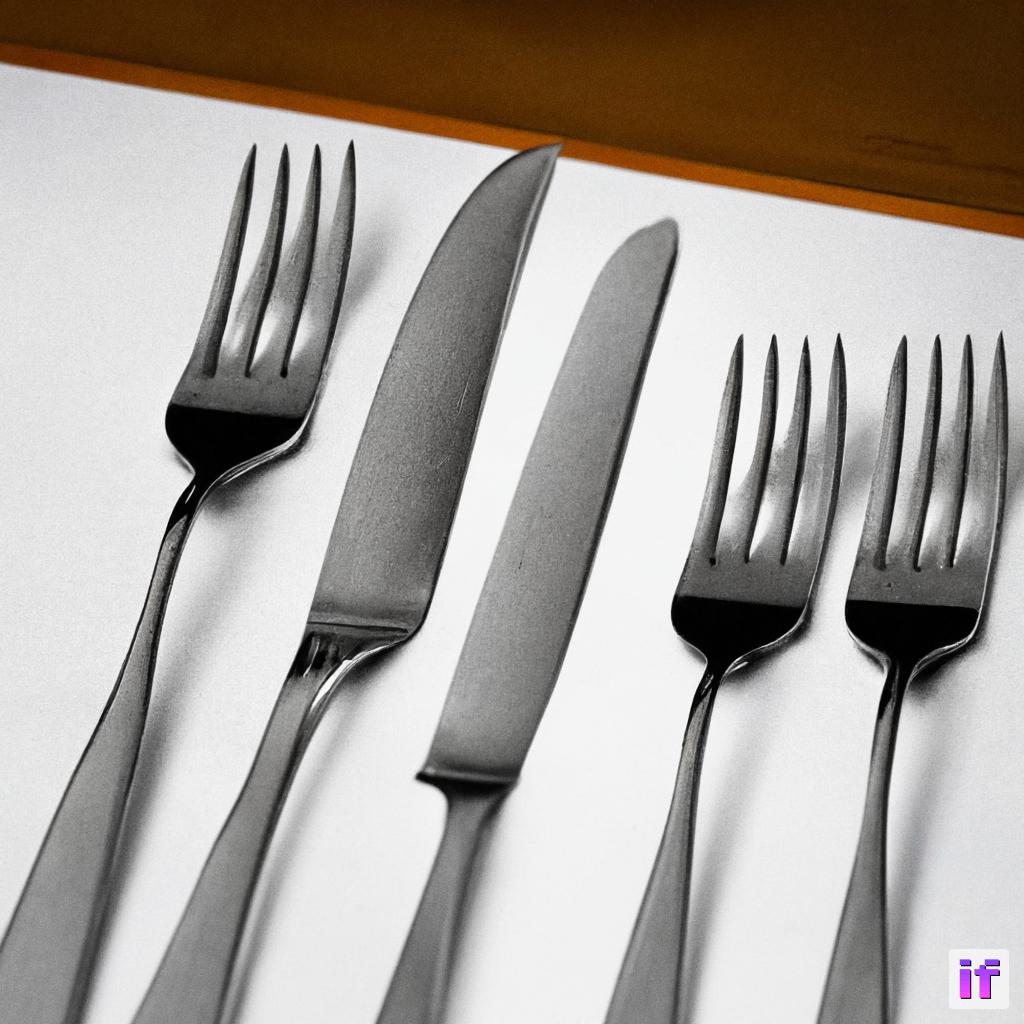}
        };

        \node(iter1)[draw,dashed,rounded corners,below=of iter0]{
            \parbox{46em}{\raggedright \textit{iteration 2:} a set of knives and forks leaning against each other on a table \textbf{\antiqueone with a white tablecloth and a vase of flowers in the background}\\ \textbf{Score: 0.939}}
        };
        \node[right=of iter1] (img1) {
            \includegraphics[width=4.5em]{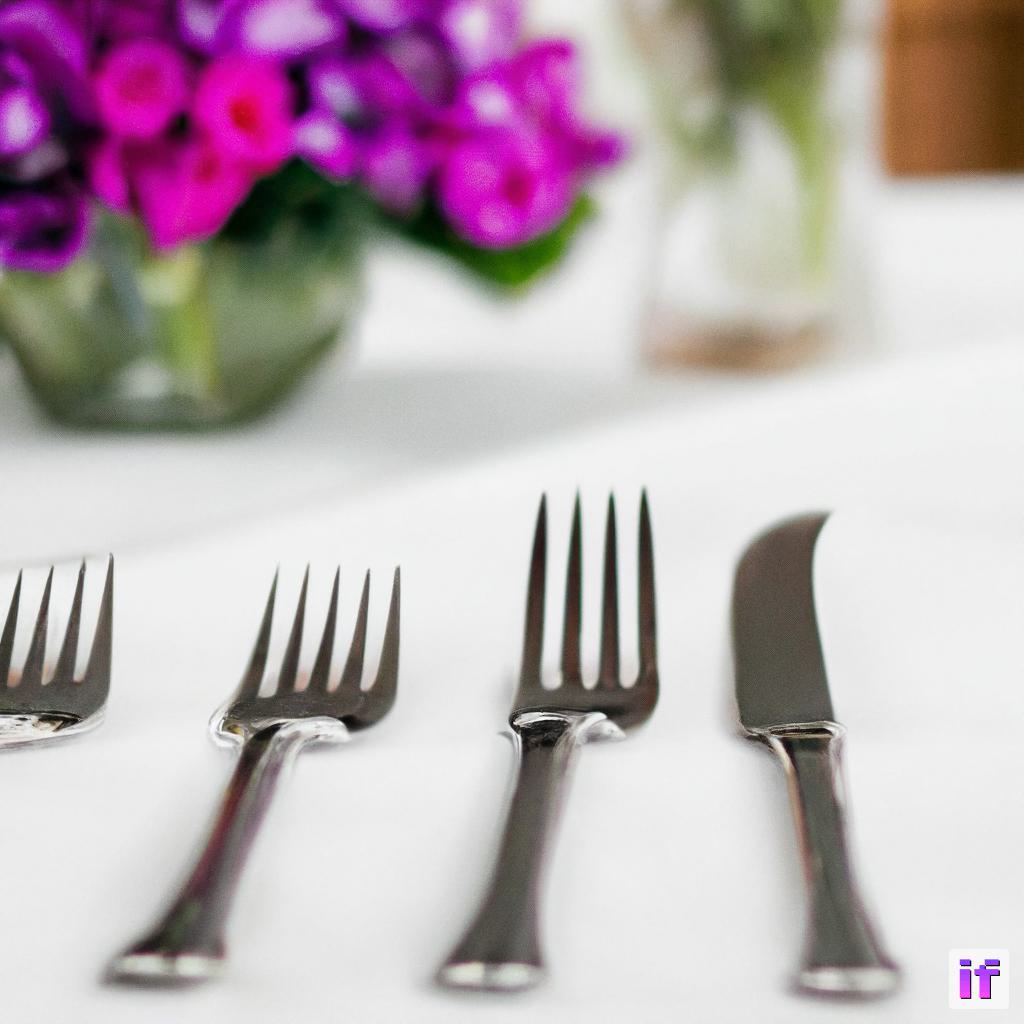}
        };

        \node(iter2)[draw,dashed,rounded corners,below=of iter1]{
            \parbox{46em}{\raggedright\textit{iteration 3:} a set of knives and forks leaning against each other on a table with a white tablecloth and a vase of flowers in the background \textbf{\antiqueone with a small wooden bowl of fruit on the table and a cat sitting on the floor next to the table}\\ \textbf{Score: 0.835}}
        };
        \node[right=of iter2] (img2) {
            \includegraphics[width=4.5em]{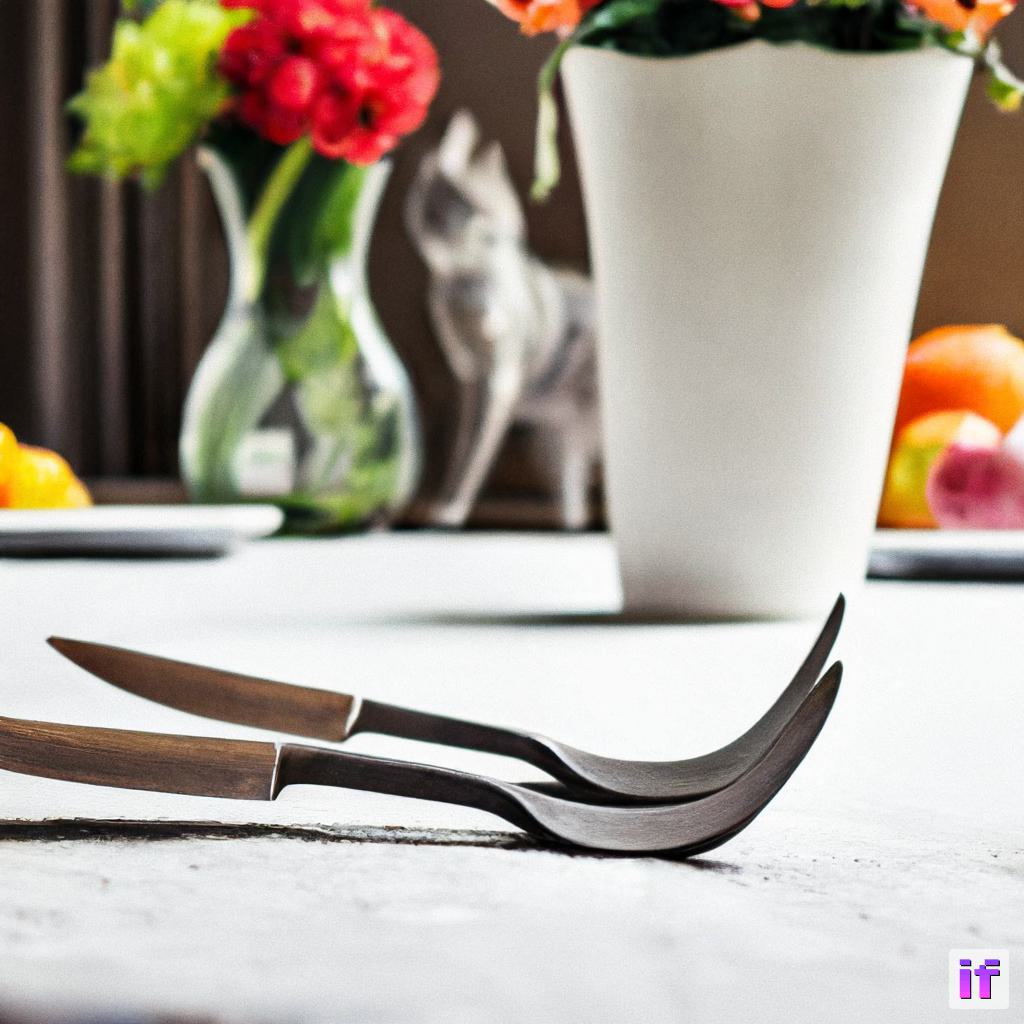}
        };

        \node(iter3)[draw,dashed,rounded corners,below=of iter2]{
            \parbox{46em}{\raggedright\textit{iteration 4:} a set of knives and forks leaning against each other on a table with a white tablecloth and a vase of flowers in the background with a small wooden bowl of fruit on the table and a cat sitting on the floor next to the table \textbf{\antiqueone and a bird perched on the tablecloth}\\ \textbf{Score: 0.386}}
        };
        \node[right=of iter3] (img3) {
            \includegraphics[width=4.5em]{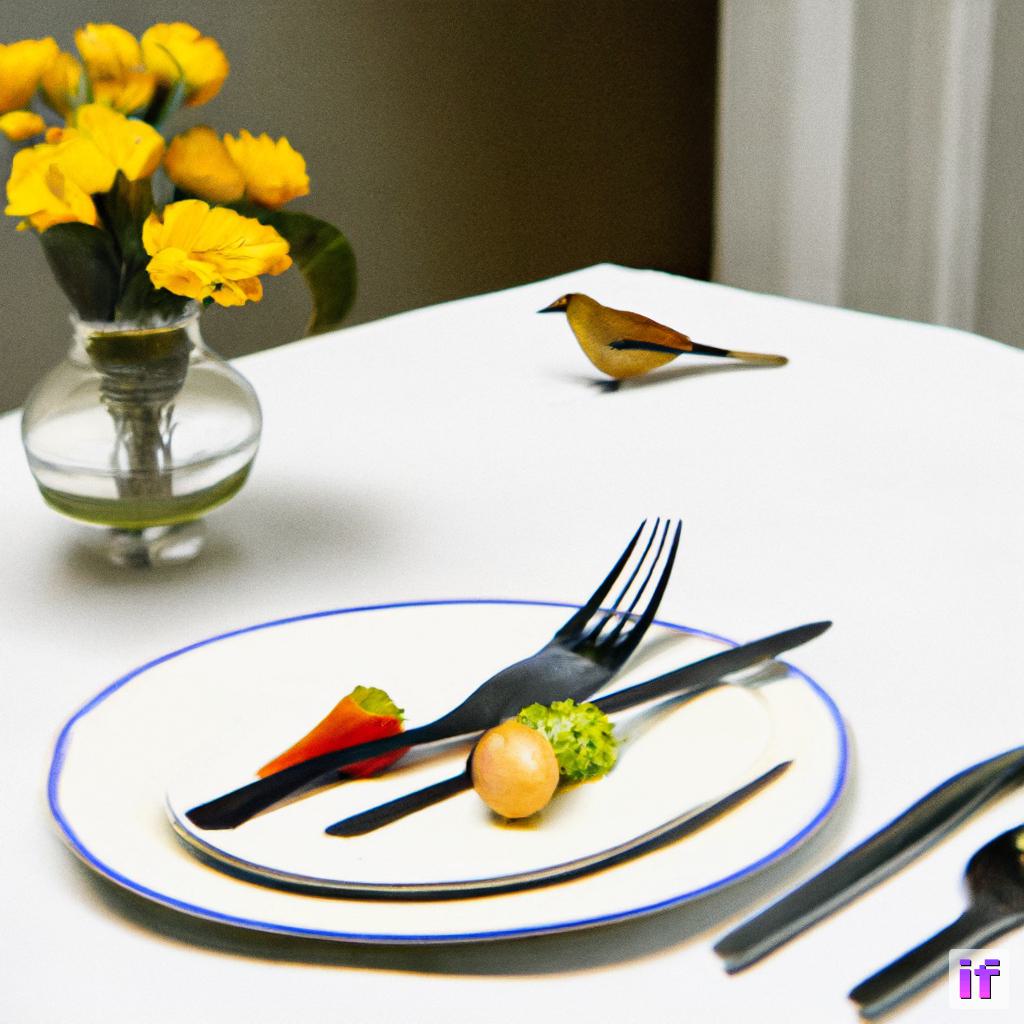}
        };

        \node(iter4)[draw,dashed,rounded corners,below=of iter3]{
            \parbox{46em}{\raggedright\textit{iteration 5:} a set of knives and forks leaning against each other on a table with a white tablecloth and a vase of flowers in the background \textbf{\removalRed [with a small wooden bowl of fruit on the table and a cat sitting on the floor next to the table and a bird perched on the tablecloth]}\\ \textbf{Score: 0.911}}
        };
        \node[right=of iter4] (img4) {
            \includegraphics[width=4.5em]{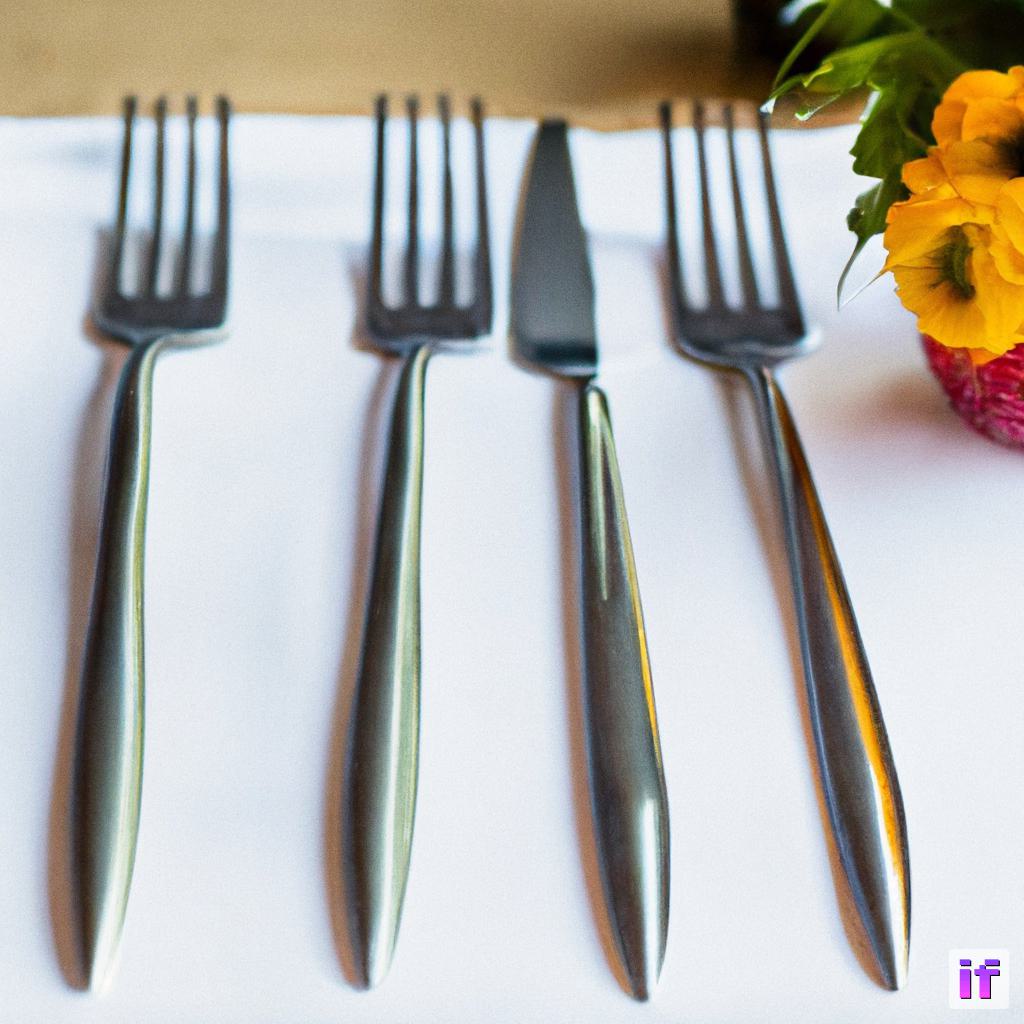}
        };

        \draw[->,line width=0.05mm] (iter0) -- (iter1);
        \draw[->,line width=0.05mm] (iter1) -- (iter2);
        \draw[->,line width=0.05mm] (iter2) -- (iter3);
        \draw[->,line width=0.05mm] (iter3) -- (iter4);

    \end{tikzpicture}
    \vspace{0.1cm}
    \caption{Example run with \method's Adaptive Evaluation, with generated images and alignment scores to illustrate the evaluation process. {\antiqueone \textbf{Portions in purple}} were added to the prompt from the previous iteration; {\removalRed \textbf{portions in red}} were removed. Generations have high scores for iterations 1-3, so the prompts get progressively more difficult. When the score drops significantly at iteration 4, \method\ reacts and simplifies the prompt.\looseness-1
    }
    \label{fig:ex-adaptive-prompts}
\end{figure}

We demonstrated that the \method\ framework can iteratively generate prompts of increasing complexity, providing more efficient benchmarks compared to traditional static prompt sets.
In this section, we move a step further in benchmark generation and ask: \textbf{Can \method\ customize evaluations for individual T2I models based on their performance history?}
Individualized evaluation provides a nuanced understanding of differences between models and can facilitate the automatic discovery of model failure points.
We explore adding an adaptive aspect to \method, where the MLLM interacts with the T2I model by adjusting prompts based on the generated images seen from previous iterations.
\vspace{-0.2cm}

\subsection{Adaptive Prompt Generation}

\begin{figure}[t]
    \centering
    \includegraphics[width=\linewidth]{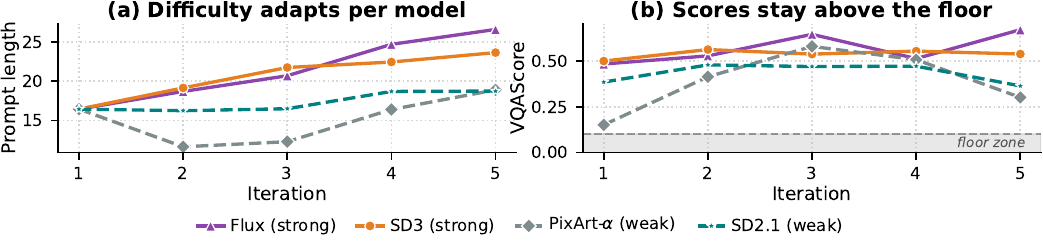}
    \vspace{-0.5cm}
    \caption{\method\ in the adaptive setting (Qwen3-VL-32B scorer), averaged over 5 seeds and 4 seed-prompt trajectories, shown for two strong (Flux, SD3; solid) and two weak (PixArt-$\alpha$, SD2.1; dashed) T2I models. \textit{(a)} Shows that prompts for strong models get progressively longer and more complex, while prompts for weak models are simplified and kept relatively shorter;
    \textit{(b)} Shows that VQAScores for all models stay in an informative range above the near-zero \emph{floor zone} (shaded) because prompt difficulty adapts per model. This fixes the issue with \method's progressive difficulty setting, where the same difficulty progression drives most models into the floor zone as prompts become more complex (see Figure~\ref{fig:adaptive_iter_ablation}a).}
    \label{fig:adaptive_mechanism}
    \vspace{-0.3cm}
\end{figure}

As previously described, \method\ iteratively expands upon prompts and scores corresponding image generations.
In this adaptive setting, the MLLM additionally incorporates feedback about how well the T2I model performed on the previous iteration, via the score (Figure \ref{fig:method}, right). This approach enables the MLLM to adaptively generate more challenging prompts when the T2I model performs well and simpler prompts when the model struggles.
This dynamic adjustment is desirable because it enables discovery of failure patterns in strong models while avoiding uninformative results from excessively difficult prompts for weak models.\looseness-1

In this setting, the MLLM is instructed to update prompt $T_{i}$ based on the image-text consistency score of the previous prompt $T_{i-1}$ according to:

\vspace{-0.3cm}
\begin{itemize}
    \setlength{\itemsep}{1pt} 
    \setlength{\parskip}{1pt} 
    \item score $0.0-0.2$: complexity halved
    \item score $0.2-0.4$: complexity reduced by 1-2 terms
    \item score $0.4-0.6$: prompt rephrased, same complexity
    \item score $0.6-0.8$: complexity increased by 1-2 terms
    \item score $0.8-1.0$: complexity increased by 2+ terms
\end{itemize}
\vspace{-0.3cm}

Figure \ref{fig:adaptive_mechanism}(a) shows how prompt difficulty shifts across iterations for two strong T2I models (Flux, SD3) and two weak ones (PixArt-$\alpha$, SD2.1). The two clusters separate over iterations: the loop escalates the strong models to progressively longer, more complex prompts, while holding the weak models to markedly shorter ones.
Correspondingly, Figure \ref{fig:adaptive_mechanism}(b) shows that, because difficulty adapts per model, all models' consistency scores stay in an informative range above the near-zero floor across the default five iterations, rather than collapsing as they do under the progressive difficulty setting (Table~\ref{tab:iter_ablation}); Section~\ref{sec:adaptive_ablation} shows this holds out to 48 prompts.
An example of an adaptive run for DeepFloyd \citep{deepfloyd} is shown in Figure \ref{fig:ex-adaptive-prompts}.

\subsection{Adaptive Prompting Preserves T2I Model Rankings}
Adaptive prompting evaluates each T2I model on its own bespoke prompt trajectory. We ask whether it can still recover the official benchmark ranking, using Experiment~2's protocol unchanged: both MLLM scales, the 8 T2I models, 5 seeds, and per-draw Kendall $\tau$ against the same official benchmark rankings. Results are therefore directly comparable to the progressive setting, and are reported alongside it in the lower blocks of Table~\ref{tab:ranking_corr}.
Since the adaptively generated prompts differ across models, a raw mean VQAScore is not comparable between them: a weak model may score highly simply because it was fed easier prompts. We therefore rank each model by a \emph{difficulty-weighted} mean consistency, weighting each score by the length (word count) of the prompt it was earned on:
\begin{equation}
\text{score}(\text{model}) = \frac{\sum_{i} \text{VQAScore}_i \cdot w_i}{\sum_{i} w_i},
\end{equation}
where $w_i$ is the word count of prompt $i$, summed over all of a model's adaptive prompts. This credits sustained consistency on the harder, longer prompts the loop escalates to, and discounts high scores earned on prompts it simplified.
Appendix~\ref{app:prompt_complexity} compares candidate difficulty metrics and shows T2I evaluation is insensitive to which length-based statistic supplies the weight; Appendix~\ref{app:escalation_depth} shows that prompt length alone, with scores discarded, is too noisy to rank by.\looseness-1

Adaptive \method\ produces positive correlations with the official rankings of all three benchmarks at both MLLM scales ($\tau=0.51$--$0.63$; Table~\ref{tab:ranking_corr}), despite evaluating every model on a \emph{different}, model-specific set of only 20 prompts.
Against the baseline metrics it wins on GenEval and GenAIBench at both scales, and on GenAIBench the 32B scorer achieves the strongest correlation overall.
T2I-CompBench is the one benchmark where a baseline remains ahead. The central result is that bespoke, per-model evaluation stays competitive with methods that score every model on the same shared prompt set.

Against \method's progressive setting, bespoke prompts do cost evaluation fidelity: adaptive trails on GenEval at both scales and leads at only one scale each on T2I-CompBench and GenAIBench.
Adaptive \method\ also incurs higher variance, with higher per-draw standard deviation across all evaluation settings.
In exchange it probes each model at its own performance frontier, keeping every model's scores in an informative range as the prompt budget grows (Section~\ref{sec:adaptive_ablation}).
The orderings behind these correlations (Figure~\ref{fig:adaptive_ranking}, Appendix~\ref{app:adaptive_ranking}) show what the lost fidelity costs in practice: the adaptive ranking still places SD3 and Flux at the top and SD2.1 and PixArt-$\alpha$ at the bottom, reordering only within those pairs, while mid-order models shift by up to two ranks.

\subsection{Adaptation Sustains Signal as Prompts Grow}\label{sec:adaptive_ablation}
By 48 prompts, $86\%$ of the progressive difficulty setting's scores sit at the floor (Section~\ref{sec:iter_ablation_body}), so most prompts are too hard only and elicit failed generations.
Adaptive \method\ is designed precisely to avoid this: by reducing prompt difficulty whenever a model starts to fail, it should keep each model in its informative range regardless of prompt iteration count.
We test this directly by running adaptive \method\ for 12 iterations (48 bespoke prompts per model) and tracking, at each prompt count, the floor fraction and the difficulty-weighted ranking correlation $\bar{\tau}$ (averaged over the three benchmarks' official rankings).

\begin{figure}[t]
    \centering
    \includegraphics[width=\linewidth]{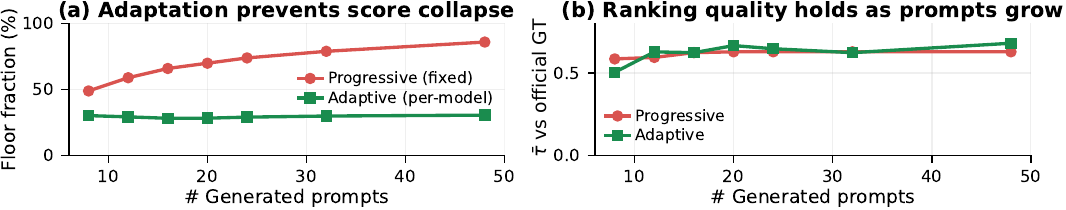}
    \caption{\textbf{Adaptation sustains signal as the prompt budget grows} (Qwen3-VL-32B scorer). \textit{(a)} Floor fraction (share of model--prompt VQAScores $\le\!0.1$) vs.\ number of generated prompts: the fixed progressively difficult prompts (Section~\ref{sec:mllm_genbench}) drives the floor from $49\%$ to $86\%$, while adaptive \method\ holds it flat at $\sim\!30\%$ by reducing difficulty per model. \textit{(b)} Ranking correlation $\bar{\tau}$ with the official benchmark rankings (averaged over the three benchmarks): both settings hold their correlation across this range, but only adaptive does so without driving its scores to the floor (a).}
    \label{fig:adaptive_iter_ablation}
    \vspace{-0.3cm}
\end{figure}

Figure~\ref{fig:adaptive_iter_ablation} confirms the adaptive design works. For progressive difficulty \method\ the floor fraction rises steeply, but under adaptive \method\ it stays \emph{flat at $\sim\!30\%$} across all iteration counts (Figure~\ref{fig:adaptive_iter_ablation}a): the loop never lets the population of models collapse toward zero.
Ranking quality holds alongside it. Adaptive \method's $\bar{\tau}$ shows no decay out to 48 prompts (Qwen3-VL-32B: $0.51$ at 8 prompts and $0.68$ at 48, a gap within run-to-run noise). Extending the loop therefore neither degrades the ranking nor drives models to the floor.
Appendix~\ref{app:adaptive_iter_ablation} gives the full per-count ablation for both MLLM scorers.
Overall, adaptive \method\ trades some ranking fidelity for per-model customization while still recovering benchmark orderings from only 20 prompts, and---unlike the fixed-schedule setting---keeps every model in an informative scoring range however far the loop is run, enabling the individualized, failure-seeking evaluation that motivated this setting.

\section{Conclusion}
We introduced \method, a framework that reframes text-to-image (T2I) evaluation as an interaction between a multimodal LLM (MLLM) evaluator agent and the T2I model under test, with the MLLM both generating the evaluation prompts and scoring the resulting images.

We established this in three experiments.
First, the MLLM can \emph{judge}: on TIFA160, Qwen3-VL-32B correlates with human ratings better than every prior automatic metric and multi-model GQA pipeline, without their separate question-generation, filtering, and VQA models.
 Second, it can \emph{generate} the benchmark. From 20 generated prompts---$28$--$105\times$ fewer than the benchmarks' own prompt sets---\method\ recovers each benchmark's official ranking of 8 T2I models, with Qwen3-VL-32B exceeding every baseline's rank correlation on all three at half their run-to-run standard deviation.
Third, it can \emph{adapt}: setting each prompt's difficulty from the T2I model's own measured performance produces a bespoke per-model trajectory whose rankings stay competitive with the strongest baselines.

Together these results address the prompt curation cost and performance saturation that plague static benchmarks~\citep{ott2022mapping,gupta2024improving, lin2014microsoft, yu2022scaling}: \method\ requires no fine-tuning, works with any off-the-shelf MLLM, and adds no cost beyond standard inference.
We detail \method's MLLM system prompts in Appendix~\ref{app:prompt_gen_meta_prompts}, its computational cost in Appendix~\ref{app:comp_cost}, and its limitations and directions for future work in Appendix~\ref{app:discussion}. We hope these findings motivate a shift from static prompt sets toward dynamic, interactive, agent-based evaluation of generative models.

\clearpage
\newpage
\bibliographystyle{assets/plainnat}
\bibliography{main}

\clearpage
\newpage
\beginappendix


\section{MT2IE For Aesthetic Quality Evaluations}\label{app:aesthetic_evals}

To demonstrate {\method}'s ability to generalize beyond evaluating image-text consistency, we use it to evaluate image aesthetic quality and compare its aesthetic quality judgments to a state-of-the-art aesthetics scoring model. \method\ is easily adaptable to be used to evaluate other aspects of generated images by simply using a different set of MLLM system prompts; for aesthetic quality evaluation we use the system prompts in Figure \ref{fig:aesthetic-prompt}.
All other aspects of \method\ are kept the same: we score with Qwen3-VL-32B, the same MLLM used in Section~\ref{sec:efficiency}, over the 100 generated prompts and 800 corresponding images (100 per T2I model) from that section.
We compare aesthetic scores produced by \method\ and the aesthetics predictor of~\citet{laion}, a popular aesthetics-scoring model \citep{pickscore, human_pref_score, imagereward}.
\method's ranking of 8 T2I models by average aesthetic score closely matches the pre-trained aesthetics-scoring model's ranking, with a \emph{Kendall's $\tau$ of $0.786$}: of the 28 model pairs, only 3 are ordered differently, with \method\ placing Flux two positions higher and swapping the two lowest-ranked models. Agreement is weaker at the level of individual images (\emph{Kendall's $\tau$ of $0.319$} over the 800 images), which is unsurprising given the resolution of the two scores: \method\ emits integers on a $0$--$10$ scale and assigns $94\%$ of images a $7$, $8$, or $9$, whereas the pre-trained predictor is continuous. \method\ therefore recovers the aggregate aesthetic ordering of T2I models considerably more reliably than it reproduces per-image aesthetic judgments.
These results, in tandem with our main paper experiments, show that \method\ can effectively evaluate aesthetic quality and image-text consistency. Additionally, we demonstrate that \method\ can be easily repurposed to evaluate axes other than image-text consistency by rewriting the MLLM prompt used for judgement.

\section{Ease of Adoption \& Computational Cost}\label{app:comp_cost}
The computational cost of one evaluation run with \method\ is $2P$ inference passes of the utilized MLLM, where $P$ is the number of generated prompts ($P$ prompt generations and $P$ prompt-image judgments); with the small prompt count we recommend ($P\!=\!20$; Table~\ref{tab:iter_ablation}), this is 40 passes.
Exact time and memory costs depend on MLLM size.
However \method\ greatly reduces the computational cost of generating images, as it recovers T2I model rankings while evaluating on far fewer image-text pairs (Section~\ref{sec:efficiency}), which offsets the MLLM computation costs.
MLLM system prompts for \method\ are provided in Appendix \ref{app:prompt_gen_meta_prompts}, \ref{app:vqa_acc_with_prompts}, and can be easily used with any off-the-shelf MLLM.

\section{GQA Scoring Results} \label{app:vqa_acc_with_prompts}

In Section \ref{sec:contribution_1}, we validate that MLLMs are able to effectively score T2I models. We have two scoring paradigms: GQA (question generation and answering) and VQAScore. GQA requires generating multiple choice questions given a description. In the main paper, we only discuss VQAScore image-text consistency scoring as it achieved the highest correlation with human judgment in experiments in Section \ref{sec:contribution_1}. We therefore provide more details on the GQA process below; the per-MLLM question generation and validation system prompts are given in Appendix~\ref{app:prompt_gen_meta_prompts}.

Figure~\ref{fig:coco_vqa} reports each MLLM's VQA accuracy on the $2{,}000$ real COCO image--caption pairs of TIFAv1.0, answered against that benchmark's original GPT-3-generated question set. This verifies that the MLLMs answer such questions accurately on real images, before we rely on them to both generate and answer questions about generated images.

\begin{figure}[]
    \centering
    \includegraphics[width=0.5\linewidth]{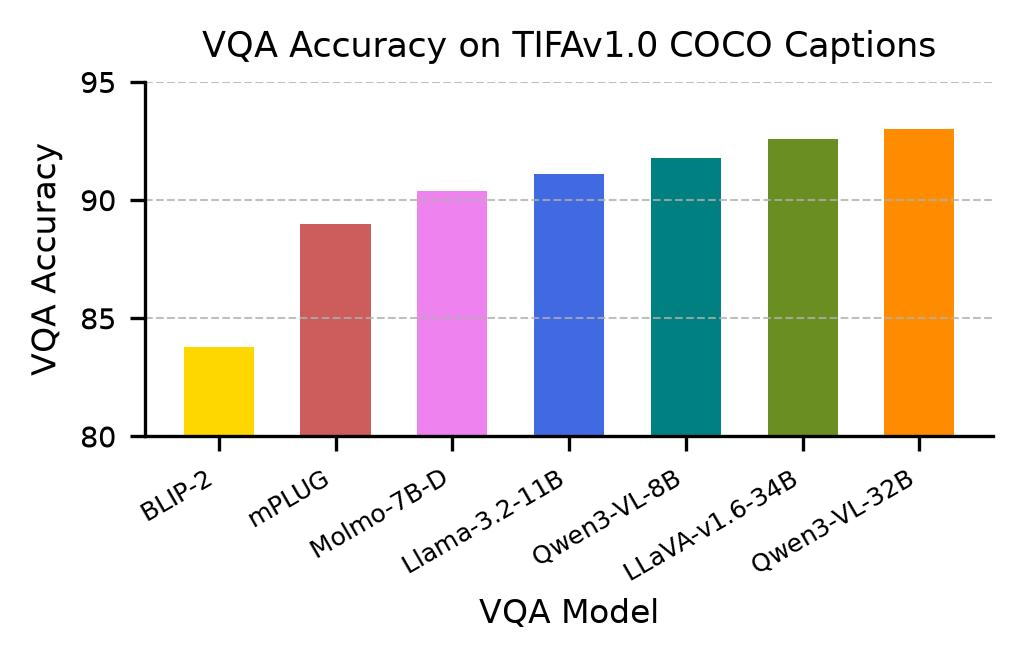}
    \caption{All MLLMs obtain higher VQA accuracy than VQA models used as image judges in previous work. Results are on the 2k COCO image-caption pairs in TIFAv1.0, as done in \citet{tifa}.
    }
    \label{fig:coco_vqa}
\end{figure}

\section{Analysis of Prompt Difficulty Metrics} \label{app:prompt_complexity}
For \method\ with Adaptive Prompting (Section \ref{sec:adaptive}), each T2I model is evaluated on its own bespoke prompt trajectory, so raw alignment scores are not directly comparable across models. We standardize scores by a proxy for prompt difficulty---weighting up scores earned on hard prompts and down those earned on easy ones. An ideal prompt difficulty metric would reflect how hard it is for a T2I model to generate an image faithful to the prompt.

To our knowledge, there is no existing work on metrics for T2I prompt difficulty, so we evaluate several NLP language-complexity statistics as proxies---word count, syllable count, and Flesch--Kincaid grade level \citep{Flesch1948ANR}---together with two per-token averages, average word length and average syllables per word, as controls. We measure how well each predicts difficulty over both the $3{,}840$ prompt--VQAScore pairs from our progressive-difficulty runs and the $1{,}600$ pairs from the adaptive runs, where the weight is actually applied.

Table \ref{tab:prompt_difficulty_corr} reports each metric's correlation with T2I VQAScore in both settings. \textbf{Only length-based metrics track difficulty}, and the three are statistically indistinguishable from one another in either setting; longer, more compositionally loaded prompts reliably yield lower consistency scores, i.e.\ are harder. The per-token averages we include as controls carry no signal in the adaptive setting, confirming that difficulty here is a matter of how many concepts a prompt asks the model to render rather than how sophisticated its vocabulary is. We weight by word count, the simplest of the three and the only one that is exactly reproducible---syllable counts depend on the syllabifier used.

We stress that word count predicts difficulty here \emph{primarily because the adaptive loop builds difficulty by adding compositional elements one at a time}, so longer prompts carry more objects, attributes, and relations to satisfy. Controlling for this---measuring the word-count/VQAScore correlation \emph{within} a fixed iteration, where length variation no longer reflects escalation depth---the correlation is substantially weaker at every iteration than the $-0.42$ pooled value, ranging from $\tau=-0.03$ at the first iteration to $-0.30$ at the tenth. The same effect explains why the correlations are weaker in the adaptive setting ($\tau=-0.28$ vs.\ $-0.42$): the adaptive loop shortens prompts whenever a model starts to fail, so length and escalation depth are less tightly coupled there. Word count is therefore best understood as a monotone proxy for \emph{how far the loop escalated a prompt}, which is exactly the difficulty axis relevant to our adaptive ranking, rather than a benchmark-agnostic measure of intrinsic prompt hardness. The choice among length-based statistics matters little: weighting by word count, syllable count, or Flesch--Kincaid grade gives mean Kendall $\tau$ of $0.567$, $0.583$, and $0.536$ across the six adaptive benchmark--scale cells of Table~\ref{tab:ranking_corr}, all well within run-to-run variation. Developing a genuinely benchmark-agnostic, predictive T2I prompt-difficulty metric remains valuable future work.

\begin{table}[h!]
\centering
\caption{
Correlation of candidate prompt-difficulty metrics with T2I VQAScore, over the $3{,}840$ prompt--image pairs from our Qwen3-VL progressive-difficulty runs and the $1{,}600$ pairs from the adaptive runs where the weight is applied. A stronger (more negative) correlation means longer/more-complex prompts are reliably harder, making the metric a more viable difficulty weight. Per-token averages are included as controls: they measure lexical rather than compositional complexity.
}
\begin{tabular}{@{}lcccc@{}}
\toprule[1.2pt]
 & \multicolumn{2}{c}{\textbf{Progressive}} & \multicolumn{2}{c}{\textbf{Adaptive}} \\
\cmidrule(lr){2-3} \cmidrule(lr){4-5}
 Prompt Metric & $\tau$ & $\rho$ & $\tau$ & $\rho$\\
\midrule
\multicolumn{5}{@{}l}{\textit{Length-based:}} \\
Word Count & $-0.421$ & $-0.439$ & $-0.280$ & $-0.403$ \\
Syllable Count & $-0.419$ & $-0.445$ & $-0.285$ & $-0.412$ \\
Flesch-Kincaid Grade Level & $-0.419$ & $-0.440$ & $-0.253$ & $-0.370$ \\
\midrule
\multicolumn{5}{@{}l}{\textit{Per-token averages (controls):}} \\
Average Word Length & $+0.171$ & $+0.212$ & $-0.048$ & $-0.073$ \\
Average Syllables per Word & $+0.148$ & $+0.138$ & $+0.001$ & $+0.001$ \\
\bottomrule[1.2pt]
\end{tabular}
\label{tab:prompt_difficulty_corr}
\end{table}

\subsection{Escalation Depth Alone Is Not Enough}\label{app:escalation_depth}
Since word count acts largely as a proxy for how far the loop escalated a prompt, a natural question is whether that escalation depth is enough on its own: can we rank T2I models by mean final prompt length alone, discarding score magnitude entirely? It is not. This score-independent ranking correlates only weakly with the official ordering, giving per-draw $\tau$ of $0.201$, $0.130$, and $0.144$ for Qwen3-VL-8B and $0.137$, $0.237$, and $0.123$ for Qwen3-VL-32B on GenEval, T2I-CompBench, and GenAIBench respectively. With 5 seeds and per-draw standard deviations of $0.15$--$0.26$, only one of these six values is distinguishable from zero at the $5\%$ level, whereas all six difficulty-weighted values in Table~\ref{tab:ranking_corr} are. How far the loop escalates a model is therefore too noisy to rank by alone, and the difficulty-weighted metric of Section~\ref{sec:adaptive} needs both of its ingredients: the scores supply the signal, and the prompt lengths make scores earned on unequal prompts comparable.

\section{Full Prompt-Count Ablation}\label{app:iter_ablation}
Table~\ref{tab:iter_ablation} in Section~\ref{sec:mllm_genbench} reports a readable subset of prompt counts. Table~\ref{tab:iter_ablation_full} gives the complete sweep over all difficulty-iteration counts (2--12 iterations, i.e.\ 8--48 generated prompts), per benchmark and per scorer, together with the score spread and floor fraction defined in Section~\ref{sec:iter_ablation_body}. The $\tau$ columns are per-draw Kendall's $\tau$ (mean over 5 seeds) against each benchmark's \emph{official benchmark ranking}, as in Table~\ref{tab:ranking_corr}; the spread and floor columns are properties of the score distribution and are independent of the ground-truth ranking.

The full sweep confirms the behavior discussed in Section~\ref{sec:mllm_genbench}: Qwen3-VL-32B rises to 20 prompts and is then essentially flat from 20 through 48 prompts on all three benchmarks (identical on GenEval and T2I-CompBench; GenAIBench peaks slightly earlier at 16), while the smaller Qwen3-VL-8B is noisier and never reaches the larger scorer's level. Across both scorers, the score spread contracts monotonically and the floor fraction rises monotonically as prompts are added, reaching $78\%$ (Qwen3-VL-8B) and $86\%$ (Qwen3-VL-32B) at 48 prompts---i.e.\ at high iteration counts the majority of generations are pushed to near-zero consistency, erasing the between-model differences a ranking relies on. This is why adding prompts beyond $\sim\!20$ yields no gain.

\begin{table}[h]
\footnotesize
\centering
\setlength{\tabcolsep}{3.5pt}
\caption{
Full prompt-count ablation for \method. Columns per scorer: per-draw Kendall's $\tau$ vs.\ the \emph{official benchmark ranking} of GenEval (GE), T2I-CompBench (CB), and GenAIBench (GAI); score spread (Spr, max$-$min mean VQAScore across the 8 T2I models); and floor fraction (Flr, share of model--prompt scores $\le\!0.1$). $\tau$ values are means over 5 seeds; spread and floor are ground-truth-independent. \textbf{Bold} marks the earliest prompt count reaching each scorer's best $\tau$ per benchmark (the larger scorer plateaus from that count onward).
}
\begin{tabular}{@{}cc ccccc c ccccc@{}}
\toprule[1.0pt]
& & \multicolumn{5}{c}{\textbf{Qwen3-VL-8B}} & & \multicolumn{5}{c}{\textbf{Qwen3-VL-32B}} \\
\cmidrule(lr){3-7} \cmidrule(lr){9-13}
\textbf{Iters} & \textbf{Prompts} & GE & CB & GAI & Spr & Flr & & GE & CB & GAI & Spr & Flr \\
\midrule
2  & 8  & $0.557$ & $0.571$ & $0.514$ & $0.25$ & $41\%$ & & $0.586$ & $0.629$ & $0.543$ & $0.36$ & $49\%$ \\
3  & 12 & $0.514$ & $0.529$ & $0.471$ & $0.26$ & $46\%$ & & $0.586$ & $0.629$ & $0.571$ & $0.35$ & $59\%$ \\
4  & 16 & $0.600$ & $0.557$ & $0.557$ & $0.24$ & $51\%$ & & $0.614$ & $0.657$ & $\mathbf{0.600}$ & $0.31$ & $66\%$ \\
5  & 20 & $0.614$ & $0.571$ & $0.571$ & $0.22$ & $57\%$ & & $\mathbf{0.629}$ & $\mathbf{0.671}$ & $0.586$ & $0.29$ & $70\%$ \\
6  & 24 & $0.600$ & $\mathbf{0.586}$ & $0.557$ & $0.21$ & $62\%$ & & $0.629$ & $0.671$ & $0.586$ & $0.26$ & $74\%$ \\
7  & 28 & $0.614$ & $0.571$ & $0.571$ & $0.18$ & $65\%$ & & $0.629$ & $0.671$ & $0.586$ & $0.23$ & $77\%$ \\
8  & 32 & $\mathbf{0.629}$ & $0.586$ & $\mathbf{0.586}$ & $0.17$ & $69\%$ & & $0.629$ & $0.671$ & $0.586$ & $0.21$ & $79\%$ \\
9  & 36 & $0.614$ & $0.571$ & $0.571$ & $0.16$ & $71\%$ & & $0.629$ & $0.671$ & $0.586$ & $0.19$ & $82\%$ \\
10 & 40 & $0.629$ & $0.586$ & $0.586$ & $0.15$ & $74\%$ & & $0.629$ & $0.671$ & $0.586$ & $0.17$ & $83\%$ \\
11 & 44 & $0.629$ & $0.586$ & $0.586$ & $0.14$ & $76\%$ & & $0.629$ & $0.671$ & $0.586$ & $0.16$ & $85\%$ \\
12 & 48 & $0.629$ & $0.586$ & $0.586$ & $0.13$ & $78\%$ & & $0.629$ & $0.671$ & $0.586$ & $0.15$ & $86\%$ \\
\bottomrule[1.0pt]
\end{tabular}
\label{tab:iter_ablation_full}
\end{table}

\section{T2I Model Orderings Under Both \method\ Settings}\label{app:adaptive_ranking}
\begin{figure}[h]
    \centering
    \includegraphics[width=\linewidth]{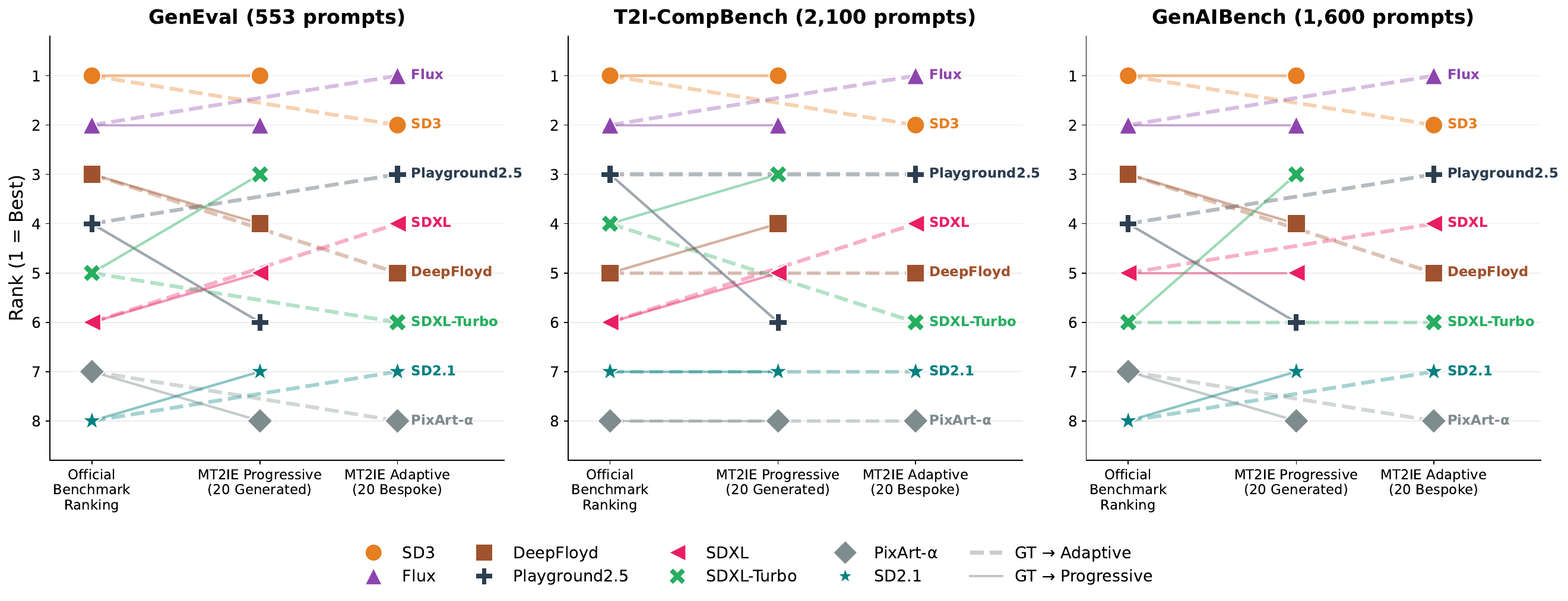}
    \caption{T2I model rankings under \method's two settings versus ground truth, across three benchmarks (Qwen3-VL-32B scorer; ground truth is each benchmark's official ranking, as in Figure~\ref{fig:all_rankings_genbench}). Each panel plots rank (1=best, 8=worst). Both lines start at the ground-truth column, so each traces one setting's deviation from the official ranking: \textbf{solid} for progressive difficulty (Section~\ref{sec:mllm_genbench}), \textbf{dashed} for adaptive; near-horizontal lines indicate a preserved ordering. Both \method\ rankings pool scores over the 5 generation seeds, as in Figure~\ref{fig:all_rankings_genbench}. The progressive ranking stays closely aligned with ground truth; the adaptive ranking preserves the top pair (SD3, Flux) and bottom pair (SD2.1, PixArt-$\alpha$) while reordering within them, and shifts mid-order models by up to two ranks.}
    \label{fig:adaptive_ranking}
\end{figure}

Table~\ref{tab:ranking_corr} summarizes each setting's ranking fidelity as a single correlation per benchmark. Figure~\ref{fig:adaptive_ranking} shows the underlying orderings, so that \emph{which} models are misranked is visible rather than only how much. Both \method\ rankings are benchmark-independent---they are computed from \method's own runs, not from any benchmark's prompts---so the two right-hand columns repeat across panels and only the ground-truth column changes.

\section{Full Adaptive Iteration Ablation}\label{app:adaptive_iter_ablation}
Section~\ref{sec:adaptive_ablation} shows, for the Qwen3-VL-32B scorer, that adaptive \method\ holds the floor fraction flat and sustains ranking quality as the prompt budget grows. Table~\ref{tab:adaptive_iter_ablation_full} gives the complete sweep over all iteration counts (2--12 iterations, i.e.\ 8--48 bespoke prompts) for both scorers. Because an adaptive trajectory depends on the scores the T2I model earns along the way, a truncated 12-iteration run does not reproduce a dedicated shorter run; this sweep therefore truncates a separate family of 12-iteration runs, and its 20-prompt entry is an independent sample of the quantity reported in Table~\ref{tab:ranking_corr} rather than the identical number ($\bar{\tau}=0.667$ here for the 32B scorer, against $0.576$ averaging that table's three 32B adaptive entries---roughly one standard error apart, given per-draw standard deviations of $0.15$--$0.24$). For each count we report the difficulty-weighted ranking correlation $\bar{\tau}$ (per-draw Kendall's $\tau$ averaged over the three benchmarks' official rankings, mean\,$\pm$\,std over 5 seeds), the score spread across the 8 T2I models, and the floor fraction. In contrast to the progressive setting (Table~\ref{tab:iter_ablation_full}), where the floor fraction rises monotonically past $85\%$, the adaptive floor fraction stays near $25$--$31\%$ at every count and both scales, and $\bar{\tau}$ shows no downward trend.

\begin{table}[h]
\footnotesize
\centering
\setlength{\tabcolsep}{4pt}
\caption{
Full adaptive iteration ablation. Per-draw $\bar{\tau}$ (difficulty-weighted ranking vs.\ the three benchmarks' official rankings, mean\,$\pm$\,std over 5 seeds), score spread, and floor fraction as the number of bespoke prompts grows. Unlike the progressive setting (Table~\ref{tab:iter_ablation_full}), the floor fraction stays flat, so ranking quality is sustained.
}
\begin{tabular}{@{}cc ccc c ccc@{}}
\toprule[1.0pt]
& & \multicolumn{3}{c}{\textbf{Qwen3-VL-8B}} & & \multicolumn{3}{c}{\textbf{Qwen3-VL-32B}} \\
\cmidrule(lr){3-5} \cmidrule(lr){7-9}
\textbf{Iters} & \textbf{Prompts} & $\bar{\tau}$ & Spread & Floor & & $\bar{\tau}$ & Spread & Floor \\
\midrule
2 & 8 & $0.290${\scriptsize$\,\pm\,0.23$} & $0.20$ & $28\%$ & & $0.505${\scriptsize$\,\pm\,0.16$} & $0.24$ & $30\%$ \\
3 & 12 & $0.419${\scriptsize$\,\pm\,0.09$} & $0.16$ & $25\%$ & & $0.629${\scriptsize$\,\pm\,0.09$} & $0.26$ & $29\%$ \\
4 & 16 & $0.486${\scriptsize$\,\pm\,0.17$} & $0.13$ & $26\%$ & & $0.624${\scriptsize$\,\pm\,0.18$} & $0.20$ & $28\%$ \\
5 & 20 & $0.505${\scriptsize$\,\pm\,0.17$} & $0.13$ & $26\%$ & & $0.667${\scriptsize$\,\pm\,0.15$} & $0.21$ & $28\%$ \\
6 & 24 & $0.495${\scriptsize$\,\pm\,0.11$} & $0.12$ & $25\%$ & & $0.648${\scriptsize$\,\pm\,0.13$} & $0.17$ & $29\%$ \\
7 & 28 & $0.543${\scriptsize$\,\pm\,0.08$} & $0.11$ & $24\%$ & & $0.624${\scriptsize$\,\pm\,0.12$} & $0.17$ & $30\%$ \\
8 & 32 & $0.552${\scriptsize$\,\pm\,0.14$} & $0.12$ & $26\%$ & & $0.624${\scriptsize$\,\pm\,0.12$} & $0.15$ & $30\%$ \\
9 & 36 & $0.471${\scriptsize$\,\pm\,0.14$} & $0.10$ & $25\%$ & & $0.738${\scriptsize$\,\pm\,0.06$} & $0.17$ & $30\%$ \\
10 & 40 & $0.562${\scriptsize$\,\pm\,0.15$} & $0.11$ & $26\%$ & & $0.610${\scriptsize$\,\pm\,0.15$} & $0.15$ & $30\%$ \\
11 & 44 & $0.543${\scriptsize$\,\pm\,0.14$} & $0.10$ & $26\%$ & & $0.695${\scriptsize$\,\pm\,0.09$} & $0.16$ & $30\%$ \\
12 & 48 & $0.529${\scriptsize$\,\pm\,0.11$} & $0.10$ & $25\%$ & & $0.681${\scriptsize$\,\pm\,0.14$} & $0.15$ & $31\%$ \\
\bottomrule[1.0pt]
\end{tabular}
\label{tab:adaptive_iter_ablation_full}
\end{table}

\section{MLLM System Prompts}\label{app:prompt_gen_meta_prompts}
The seed, progressive-difficulty, and adaptive prompt-generation system prompts below are used unchanged across every MLLM; only the MLLM being prompted differs. The question-generation prompts used for GQA scoring differ per MLLM and are given separately in Appendix~\ref{app:qgen_prompts}. All system prompts in this appendix are reproduced \emph{verbatim} from our implementation, including their original typographical errors (\eg ``givng'', ``decription''), so that the runs reported in the paper can be reproduced exactly. Seed prompts are sampled at temperature $0.5$--$0.8$ depending on topic category. Progressive-difficulty prompts are generated greedily. Adaptive prompts are sampled at temperature $0.3$ ($0.5$ for the lowest score bin) with no top-$k$ truncation.

\subsection{Seed Prompt Generation}
To generate the 4 seed prompts that begin each run, the MLLM is prompted once per topic category with the system prompt in Figure \ref{fig:seed_sys_prompt} and the user message ``\texttt{List 5 different <category>}'', where \texttt{<category>} is one of the 4 categories spanning COCO's content types (Section~\ref{sec:mllm_genbench}); one prompt is then sampled per category.

\begin{figure}[h]
\noindent\fbox{%
    \parbox{0.95\textwidth}{%
        \texttt{
        You are responsible for listing different descriptions of given topics. \\
        Each phrase should describe a few different objects and their attributes, actions, or spatial positions. \\
        You will be asked to list 5 different things each time and you must write a numbered list and state nothing else. \\
        Do not use subjective descriptors. \\
        Do not use emotional or creative language. \\
        Each item in the list should be unique from the other items.
        }
    }%
}
\caption{System prompt used by \method\ when generating the seed prompts that begin each run. The same prompt is used across all MLLMs.}
\label{fig:seed_sys_prompt}
\end{figure}

\subsection{Progressive Difficulty Prompt Generation}
For the progressive difficulty setting (Section~\ref{sec:mllm_genbench}), the MLLM rewrites each prompt to be more difficult using the system prompt in Figure~\ref{fig:iterative_sys_prompt}.

\begin{figure}[h]
\noindent\fbox{%
    \parbox{0.95\textwidth}{%
        \texttt{
        You are responsible for rewriting prompts for a computer program that generates images. \\
        You will be given an existing prompt that you must add complexity to by adding one term that is an additional object, a new spatial relationship, or new attribute.\\
        Do not add subjective descriptors to the prompt, it should be as factual as possible. Do not use fluffy, poetic language, or any words beyond the elementary school level. \\
        Every time you will be givng an existing prompt, write "Prompt:" and write the new prompt, do not state anything else.\\
\\
        Existing prompt: "a serene mountain lake with a small wooden dock and a small wooden cabin on the shore"\\
        Prompt: a serene mountain lake with a small wooden dock and a small wooden cabin on the shore, with birds in the sky \\
\\
        Existing prompt: "elderly man wearing a backpack and a hat walking with a cane and a dog on a leash"\\
        Prompt: elderly man wearing a backpack and a hat walking with a cane and a big white dog on a leash \\
\\
        Existing prompt: "a green fish hiding in a coral reef next to a crab"\\
        Prompt: a green fish hiding in a coral reef next to a crab and a purple octopus \\
        Existing prompt: (previous prompt)
        }
    }%
}
\caption{System prompt used by \method\ when generating iteratively more difficult prompts. The same prompt is used across all MLLMs.}
\label{fig:iterative_sys_prompt}
\end{figure}

\subsection{Adaptive Prompt Generation}
For the adaptive setting (Section~\ref{sec:adaptive}), the MLLM rewrites each prompt using one of five system prompts selected by the previous prompt's consistency score bin, shown in Figures~\ref{fig:adaptive_sys_prompt_1},~\ref{fig:adaptive_sys_prompt_2}, and~\ref{fig:adaptive_sys_prompt_3}.

\begin{figure}
\noindent\fbox{%
    \parbox{0.95\textwidth}{%
    \textbf{Score $[0.0, 0.2]$:}\\
        \texttt{
        You are responsible for rewriting prompts for a computer program that generates images. \\
        You will be given an existing prompt that you must simplify so that there are half the amount of terms. \\
        Remove objects, attributes, or spatial relationships in the existing prompt and rewrite the prompt so that the new prompt makes sense but only describes half the amount of concepts. \\
        If the existing prompt is too simple to reduce, just rewrite the prompt to be clearer. \\
        Every time you will be givng an existing prompt, write "Prompt:" and write the new prompt, do not state anything else.\\
\\
        Existing prompt: "a serene mountain lake with a small wooden dock a few birds flying overhead and a small wooden cabin on the shore"\\
        Prompt: a mountain lake with a wooden dock \\
\\
        Existing prompt: "elderly man walking with a cane a dog on a leash a backpack on his back a hat on his head and a pair of sunglasses on his face"
        Prompt: an elderly man walking with a cane and a dog \\
\\
        Existing prompt: "green fish swimming in a coral reef next to a purple octopus"
        Prompt: a green fish in a coral reef
        }
    }}
    \noindent\fbox{%
    \parbox{0.95\textwidth}{%
        \textbf{Score $(0.2, 0.4]$:}\\
        \texttt{
        You are responsible for rewriting prompts for a computer program that generates images. \\
        You will be given an existing prompt that you must simplify by removing two terms. \\
        The terms to remove are objects, attributes, or spatial relationships in the existing prompt. \\
        If there are less than two terms in the existing prompt, just rewrite the prompt to be simpler. \\
        Every time you will be givng an existing prompt, write "Prompt:" and write the new prompt, do not state anything else.\\
\\
        Existing prompt: "a serene mountain lake with a small wooden dock a few birds flying overhead and a small wooden cabin on the shore"\\
        Prompt: a mountain lake with a small wooden dock and a few birds \\
\\
        Existing prompt: "elderly man walking with a cane a dog on a leash a backpack on his back a hat on his head and a pair of sunglasses on his face"\\
        Prompt: elderly man with a cane wearing a backpack and a dog on a leash \\
\\
        Existing prompt: "a green, striped fish hiding in a coral reef next to a crab"\\
        Prompt: a green fish in a coral reef\\
        }
        }}
\caption{System prompts used by \method\ when generating adaptive prompts, for each of the score bins listed. The same prompts are used across all MLLMs.}
\label{fig:adaptive_sys_prompt_1}
\end{figure}

\begin{figure}
\vspace{-0.3cm}
    \noindent\fbox{%
    \parbox{0.95\textwidth}{%
        \textbf{Score $(0.4, 0.6)$:}\\
        \texttt{
        You are responsible for rewriting prompts for a computer program that generates images. \\
        You will be given an existing prompt that you must rewrite to make more clear and simple.\\
        Rewrite the prompt with simpler words, clearer phrasing, and add punctuation if needed. The meaning of the new and existing prompts must be the same.\\
        Every time you will be givng an existing prompt, write "Prompt:" and write the new prompt, do not state anything else.\\
\\
        Existing prompt: "a serene mountain lake with a small wooden dock a few birds flying overhead and a small wooden cabin on the shore"\\
        Prompt: a lake on a mountain with a small dock, birds flying in the sky, and a small cabin on the shore \\
\\
        Existing prompt: "elderly man walking with a cane a dog on a leash a backpack on his back a hat on his head and a pair of sunglasses on his face"\\
        Prompt: an old man wearing a backpack, a hat, and sunglasses walking with a cane and his dog on a leash\\
\\
        Existing prompt: "a group of seals jumping out of the water in a large body of water with a boat nearby and a person on the boat taking a picture while a seagull is flying overhead"\\
        Prompt: seals jumping out of a large body of water, near a boat with a person taking a picture onboard and a seagull flying overhead \\
\\
        Existing prompt: "a city with crowded streets and skyscrapers with a mountain in the background and traffic in the foreground and a large stadium in the distance"\\
        Prompt: a city with busy streets and tall buildings and a mountain in the background and cars in the foreground, with a large stadium in the distance\\
        }
    }}
    \noindent\fbox{%
    \parbox{0.95\textwidth}{%
        \textbf{Score $[0.6, 0.8)$:}\\
        \texttt{
         You are responsible for rewriting prompts for a computer program that generates images. \\
        You will be given an existing prompt that you must add complexity to by adding one term that is an additional object, a new spatial relationship, or new attribute.\\
        Do not add subjective descriptors to the prompt, it should be as factual as possible. Do not use fluffy, poetic language, or any words beyond the elementary school level. \\
        Every time you will be givng an existing prompt, write "Prompt:" and write the new prompt, do not state anything else.\\
\\
        Existing prompt: "a serene mountain lake with a small wooden dock and a small wooden cabin on the shore"\\
        Prompt: a serene mountain lake with a small wooden dock and a small wooden cabin on the shore, with birds in the sky \\
\\
        Existing prompt: "elderly man wearing a backpack and a hat walking with a cane and a dog on a leash"\\
        Prompt: elderly man wearing a backpack and a hat walking with a cane and a big white dog on a leash \\
\\
        Existing prompt: "a green fish hiding in a coral reef next to a crab"\\
        Prompt: a green fish hiding in a coral reef next to a crab and a purple octopus
        } }}
        \vspace{-0.3cm}
\caption{System prompts used by \method\ when generating adaptive prompts, for each of the score bins listed. The same prompts are used across all MLLMs.}
\label{fig:adaptive_sys_prompt_2}
\end{figure}

\begin{figure}
    \noindent\fbox{%
    \parbox{0.95\textwidth}{%
    \textbf{Score $[0.8, 1.0]$:}\\
        \texttt{
        You are responsible for rewriting prompts for a computer program that generates images. \\
        You will be given an existing prompt that you must add complexity to by adding a few terms. Each term is an additional object, a new spatial relationship, or new attribute.\\
        Do not add subjective descriptors to the prompt, it should be as factual as possible. Do not use fluffy, poetic language, or any words beyond the elementary school level. \\
        Every time you will be givng an existing prompt, write "Prompt:" and write the new prompt, do not state anything else.\\
\\
        Existing prompt: "a serene mountain lake with a small wooden dock and a cabin on the shore"\\
        Prompt: a serene mountain lake with a small wooden dock and a large cabin on the shore, with birds in the sky and a fisherman fishing on the dock\\
\\
        Existing prompt: "elderly man wearing a backpack and a hat walking with a cane and a dog on a leash"\\
        Prompt: elderly man wearing a blue backpack, a red hat, and sunglasses walking with a cane and a big white dog on a leash \\
\\
        Existing prompt: "a green fish hiding in a coral reef next to a crab"\\
        Prompt: a striped green fish hiding in a coral reef next to a crab and a purple octopus with the open ocean in the background\\
\\
        Existing prompt: "a lemon tree on a hill"\\
        Prompt: a big lemon tree on a hill with a dog sitting underneath\\
        } }}
\caption{System prompts used by \method\ when generating adaptive prompts, for the score bin listed. The same prompts are used across all MLLMs.}
\label{fig:adaptive_sys_prompt_3}
\end{figure}

\subsection{Image Scoring (GQA Question Generation)}\label{app:qgen_prompts}
For the GQA scoring paradigm (Appendix~\ref{app:vqa_acc_with_prompts}), each MLLM generates and validates questions about a prompt using the system prompts below, which---unlike the prompt-generation prompts above---differ per MLLM: Figure~\ref{fig:llama-prompt} (Llama-3.2-11B-Vision-Instruct), Figure~\ref{fig:molmo-prompt} (Molmo-7B-D), Figure~\ref{fig:llava-prompt} (LLaVA-v1.6-34B), and Figure~\ref{fig:qwen-prompt} (Qwen3-VL-8B-Instruct and Qwen3-VL-32B-Instruct, which share one set of prompts). We only keep questions the validation prompt answers ``yes''.

\begin{figure}[h!]
    \centering

    \noindent\fbox{%
    \parbox{0.95\textwidth}{%
    \texttt{
    Given an image description, generate multiple-choice questions that check if an image depicts everything in the image description.\\
    You will not be given an image, just generate multiple-choice questions for the given image description and write nothing else.}\\
    \\
    \texttt{First list the objects, attributes, spatial relations, and actions in the image description. Then write one questions for each of the stated items.\\
    Write one question checking if each object is in the image.
    For any object attributes, write a question checking if the object in the image has the attribute.\\
    For any spatial relations or actions, write a question checking if the spatial relation or action is shown in the image. }\\
    \\
    \texttt{For each multiple-choice question, list answer choices and state the correct answer. \\
    First write "Q:" and then state the question and nothing else. \\
    Then on the next line write "Choices: " and then list the answer choices all on one line.\\
    Then on the next line write "A: " and then state which answer out of the choices is correct. \\
    All answer choices should be relevant to the question and short.}
        }%
    }
    \vspace*{1.5ex}
    \noindent\fbox{%
        \parbox{0.95\textwidth}{%
            \texttt{The only information you have is this image description: (prompt)}

            \texttt{Can this question be answered with only the image description?}

            \texttt{Question: (question)}

            \texttt{Answer yes or no and state nothing else.}
        }%
    }

    \caption{For Llama-3.2-11B-Vision-Instruct, we provide the prompt for the question generation (top) and question validation (bottom).}
    \label{fig:llama-prompt}
\end{figure}

\begin{figure}[h!]
\noindent\fbox{%
\parbox{0.95\textwidth}{%
\scriptsize\texttt{
Given an image description, generate multiple-choice questions that check if an image matches the image description. \\
You will not be given an image, just generate multiple-choice questions for the given image description and write nothing else.\\
For each multiple-choice question, generate answer choices and the correct answer. \\
Write one question for each object, attribute, and spatial relation in the image description. \\
Do not write questions about attributes of objects not mentioned in the description. Do not list more than 4 answer choices.\\
\\
Generate questions, answer choices, and answers formatted like these examples:\\
Image description: a drawing of a red crab \\
Q: Is there a crab in the image?\\
Choices: yes, no\\
A: yes\\
Q: What animal is this? \\
Choices: lobster, fish, crab, eel\\
A: crab\\
Q: Is this a drawing?\\
Choices: yes, no\\
A: yes\\
Q: What color is the crab? \\
Choices: grey, blue, red, purple\\
A: red\\
\\
Image description: A cheese grater with a red handle sitting on a white countertop next to a yellow cutting board and a green apple\\
Q: Is there a cheese grater?\\
Choices: yes, no\\
A: yes\\
Q: What kitchen tool is on the counter?\\
Choices: blender, coffee maker, cheese grater, mixer\\
A: cheese grater\\
Q: What color is the handle of the cheese grater?\\
Choices: green, silver, white, red\\
A: red\\
Q: What color is the countertop?\\
Choices: black, white, marble, grey\\
A: white\\
Q: Is there a cutting board?\\
Choices: yes, no\\
A: yes\\
Q: What color is the cutting board?\\
Choices: white, yellow, blue, brown\\
A: yellow\\
Q: Is there an apple?\\
Choices: yes, no\\
A: yes\\
Q: What color is the apple?\\
Choices: yellow, red, green\\
A: green\\
\\
Image description: laundry detergent\\
Q: Is there laundry detergent in the image? \\
Choices: yes, no\\
A: yes}

\scriptsize\texttt{Image description: a tropical rainforest with a waterfall cascading down a cliff a group of monkeys swinging from the trees and a large colorful parrot perched on a branch\\
Q: Is there a tropical rainforest?\\
Choices: yes, no\\
A: yes\\
Q: Is there a waterfall cascading down a cliff?\\
Choices: yes, no\\
A: yes\\
Q: Are there a group of monkeys in the image?\\
Choices: yes, no\\
A: yes\\
Q: What are the monkeys doing?\\
Choices: eating, swinging from trees, swimming, sleeping\\
A: swinging from trees\\
Q: Is there a parrot in the image?\\
Choices: yes, no\\
A: yes\\
Q: Where is the parrot in the image?\\
Choices: flying overhead, perched on a branch, standing on a rock, flying above the water\\
A: perched on a branch
}
}
}
\vspace*{1.5ex}
\noindent\fbox{%
    \parbox{0.95\textwidth}{%
        \scriptsize\texttt{
        Given this image caption: (prompt)\\
        Does this question ask about information explicitly included in the caption: (question)\\
        Answer yes or no and state nothing else.
        }
    }%
}
\caption{For Molmo-7B-D, we provide the prompt for the question generation (top box) and question validation (bottom box).}
\label{fig:molmo-prompt}
\end{figure}

\begin{figure}[h!]
\centering
\noindent\fbox{%
\parbox{0.95\textwidth}{%
\scriptsize\texttt{
Given an image description, generate multiple-choice questions that check if an image matches the image description. \\
You will not be given an image, just generate multiple-choice questions for the given image description and write nothing else.\\
For each multiple-choice question, generate answer choices and the correct answer. \\
Write one question for each object, attribute, and spatial relation in the image description. \\
Do not write questions about attributes of objects not mentioned in the description. Do not list more than 4 answer choices.\\
\\
Generate questions, answer choices, and answers formatted like these examples:\\
Image description: a drawing of a red crab \\
Q: Is there a crab in the image?\\
Choices: yes, no\\
A: yes\\
Q: What animal is this? \\
Choices: lobster, fish, crab, eel\\
A: crab\\
Q: Is this a drawing?\\
Choices: yes, no\\
A: yes\\
Q: What color is the crab? \\
Choices: grey, blue, red, purple\\
A: red\\
\\
Image description: A cheese grater with a red handle sitting on a white countertop next to a yellow cutting board and a green apple\\
Q: Is there a cheese grater?\\
Choices: yes, no\\
A: yes\\
Q: What kitchen tool is on the counter?\\
Choices: blender, coffee maker, cheese grater, mixer\\
A: cheese grater\\
Q: What color is the handle of the cheese grater?\\
Choices: green, silver, white, red\\
A: red\\
Q: What color is the countertop?\\
Choices: black, white, marble, grey\\
A: white\\
Q: Is there a cutting board?\\
Choices: yes, no\\
A: yes\\
Q: What color is the cutting board?\\
Choices: white, yellow, blue, brown\\
A: yellow\\
Q: Is there an apple?\\
Choices: yes, no\\
A: yes\\
Q: What color is the apple?\\
Choices: yellow, red, green\\
A: green\\
\\
Image description: laundry detergent\\
Q: Is there laundry detergent in the image? \\
Choices: yes, no\\
A: yes}
}}
\vspace*{1.5ex}
\noindent\fbox{%
    \parbox{0.95\textwidth}{%
        \scriptsize\texttt{
        Given a caption and a question, determine if the question can be answered with only the information in the caption.\\
        Answer with yes or no and write nothing else.\\
        A question can only be answered if the answer information is explicitly included in the caption.\\
        Do not allow any information to be inferred from the caption.
        }
    }%
}
\caption{For LLaVA-v1.6-34B, we provide the prompt for the question generation (top) and question validation (bottom).}
\label{fig:llava-prompt}
\end{figure}

\begin{figure}[h!]
\centering
\noindent\fbox{%
\parbox{0.95\textwidth}{%
\scriptsize\texttt{
Given an image description, generate multiple-choice questions that check if an image matches the image description.\\
The questions should cover all objects in the image description.\\
Do not write questions about attributes of objects not mentioned in the description. Do not list more than 4 answer choices.\\
Do not write questions that cannot be answered with only the image decription.\\
\\
Image description: A red colored dog.\\
Q: Is there a dog?\\
Choices: yes, no\\
Q: what animal is in the picture?\\
Choices: dog, cat, bird, fish\\
A: dog\\
Q: what color is the dog?\\
Choices: red, black, white, yellow\\
A: red\\
\\
Image description: A cheese grater with a red handle sitting on a white countertop next to a yellow cutting board and a green apple\\
Q: Is there a cheese grater?\\
Choices: yes, no\\
A: yes\\
Q: What kitchen tool is on the counter?\\
Choices: blender, coffee maker, cheese grater, mixer\\
A: cheese grater\\
Q: What color is the handle of the cheese grater?\\
Choices: green, silver, white, red\\
A: red\\
Q: What color is the countertop?\\
Choices: black, white, marble, grey\\
A: white\\
Q: Is there a cutting board?\\
Choices: yes, no\\
A: yes\\
Q: What color is the cutting board?\\
Choices: white, yellow, blue, brown\\
A: yellow\\
Q: Is there an apple?\\
Choices: yes, no\\
A: yes\\
Q: What color is the apple?\\
Choices: yellow, red, green\\
A: green\\
\\
Image description: a tropical rainforest with a waterfall cascading down a cliff a group of monkeys swinging from the trees and a large colorful parrot perched on a branch\\
Q: Is there a tropical rainforest?\\
Choices: yes, no\\
A: yes\\
Q: Is there a waterfall cascading down a cliff?\\
Choices: yes, no\\
A: yes\\
Q: Are there a group of monkeys in the image?\\
Choices: yes, no\\
A: yes\\
Q: What are the monkeys doing?\\
Choices: eating, swinging from trees, swimming, sleeping\\
A: swinging from trees\\
Q: Is there a parrot in the image?\\
Choices: yes, no\\
A: yes\\
Q: Where is the parrot in the image?\\
Choices: flying overhead, perched on a branch, standing on a rock, flying above the water\\
A: perched on a branch}
}}
\vspace*{1.5ex}
\noindent\fbox{%
    \parbox{0.95\textwidth}{%
        \scriptsize\texttt{
        Given a caption and a question, determine if the question can be answered with only the information in the caption.\\
        Answer with yes or no and write nothing else.\\
        A question can only be answered if the answer information is explicitly included in the caption.\\
        Do not allow any information to be inferred from the caption.
        }
    }%
}
\caption{For Qwen3-VL-8B-Instruct and Qwen3-VL-32B-Instruct, we provide the prompt for the question generation (top) and question validation (bottom). Both Qwen3-VL models use the same pair of prompts.}
\label{fig:qwen-prompt}
\end{figure}

\subsection{Aesthetic Quality Evaluation}
To evaluate image aesthetics (Appendix~\ref{app:aesthetic_evals}), \method\ scores each image with the system and user prompts in Figure~\ref{fig:aesthetic-prompt}.

\begin{figure}[h]
    \centering

    \noindent\fbox{%
    \parbox{0.95\textwidth}{
    \texttt{
    You are a professional image judge. You will give a score between 0 and 10 to the given images about how visually pleasing they are.
    }
    \\
    \texttt{
    Here are things to consider when scoring the images: \\
    How clear and sharp is the image? \\
    Is anything in the image malformed? \\
    Are the colors and figures in the image pleasing and coherent?\\
    }
    \\
    \texttt{
    A score of 0 indicates that the image is very blurry, only has malformed objects, and unpleasant colors. \\
    A score of 10 indicates that the image is clear, all objects are nicely depicted, and the colors are pleasant. \\
    }
        }
    }
    \vspace*{1.5ex}
    \noindent\fbox{%
        \parbox{0.95\textwidth}{%
            \texttt{Score this image between 0 and 10. Only state the score and nothing else. (image)}
        }%
    }
    \vspace{-0.3cm}
    \caption{Aesthetic quality evaluation system prompt (top) and user prompt given per-image evaluation (bottom).}
    \label{fig:aesthetic-prompt}
\end{figure}

\section{Comparing TIFAv1.0 and the LLaVA-Generated Question Set from TIFAv1.0 Prompts} \label{app:llava-tifa}

\begin{figure}[htb]
    \centering
    \includegraphics[width=0.8\textwidth]{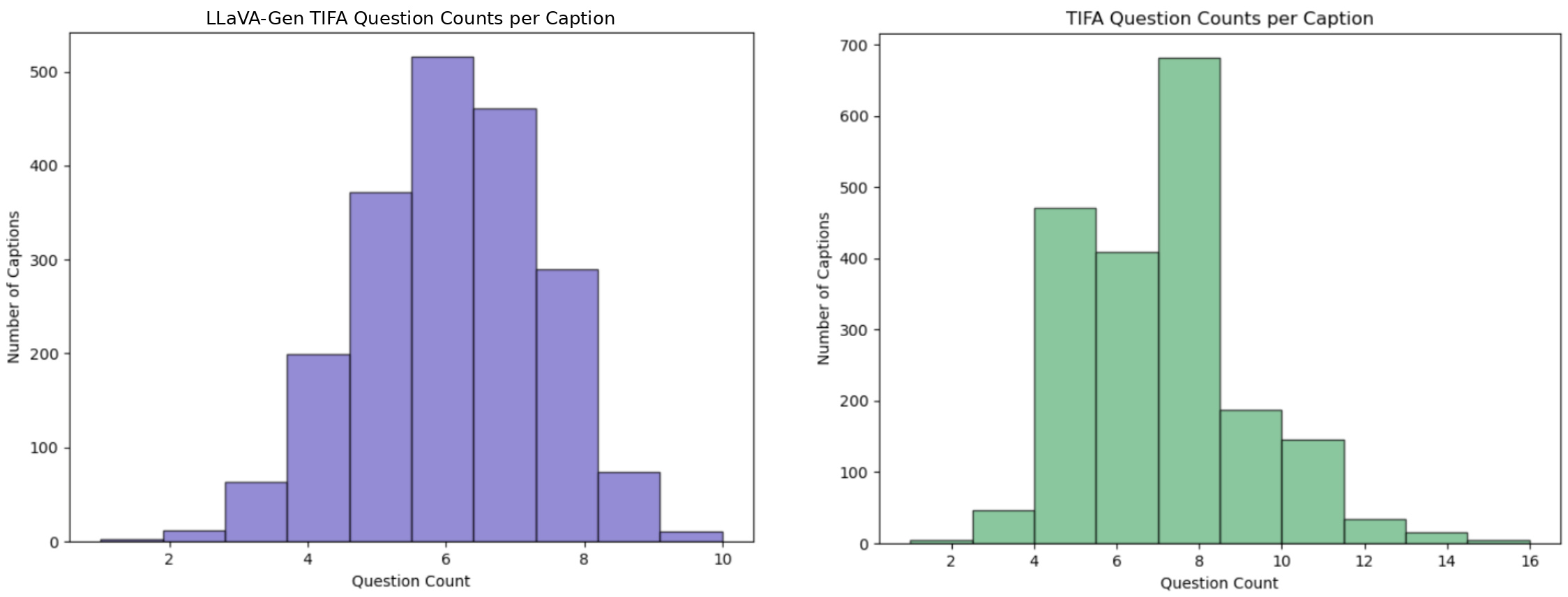}
    \caption{Distribution of the number of generated questions per caption, over all 4,081 TIFAv1.0 captions (prompts), for LLaVA-Gen TIFA (left) and TIFAv1.0 (right). The distributions are closely comparable---LLaVA-Gen TIFA averages 6.14 questions per caption versus TIFAv1.0's 6.87, with a slightly narrower tail---indicating that LLaVA decomposes prompts into question sets of similar granularity to the proprietary GPT-3. \looseness-1
    }
    \label{fig:tifa_q_counts}
\end{figure}

We analyze and compare TIFAv1.0 and our LLaVA-Generated question set (LLaVA-Gen TIFA) for the same 4,081 prompts from TIFA. This is done to verify that LLaVA is capable of generating questions as effectively as proprietary LLMs (TIFAv1.0 uses GPT-3 to generate questions).
\vspace{-0.3cm}

\paragraph{General question set statistics}
Overall, TIFAv1.0 has an average of 6.87 questions per prompt and LLaVA-Gen TIFA has an average of 6.14 questions per prompt. This results in LLaVA-Gen TIFA containing around 10\% fewer questions than TIFAv1.0 while covering the same number of prompts. Despite this, LLaVA-Gen TIFA achieves higher correlations with human judgment than TIFAv1.0, supporting our claims that MLLM-generated T2I benchmarks are more efficient.
The distribution of question counts per prompt is shown in Figure \ref{fig:tifa_q_counts}.

\vspace{-0.3cm}
\paragraph{Similarity of question elements}
TIFAv1.0 provides entity tags for each of their prompts, indicating which prompt elements the generated questions cover \citep{tifa}. We prompt LLaVA to similarly identify elements when generating questions and analyze the elements captured in TIFAv1.0 vs. LLaVA-Gen TIFA.

\begin{figure}[b]
    \centering
    \includegraphics[width=0.5\linewidth]{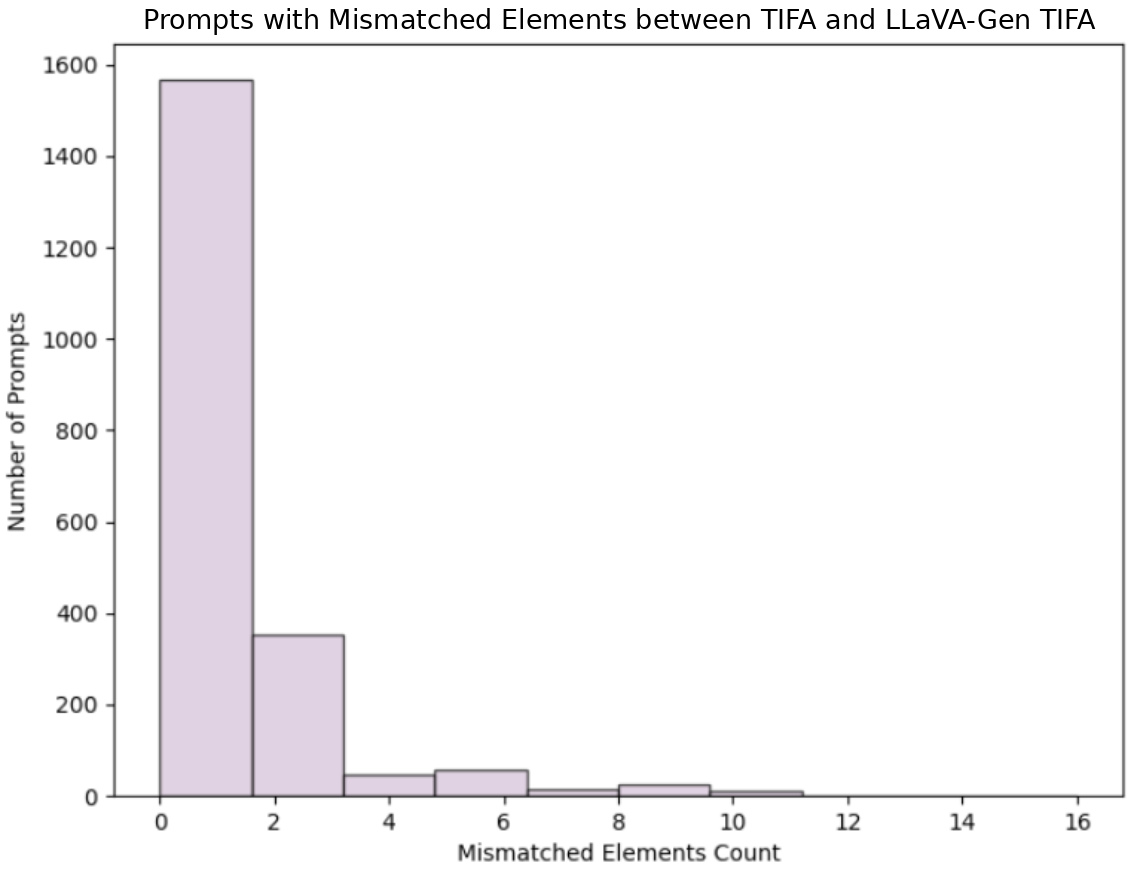}
    \vspace{-0.3cm}
    \caption{Element mismatch counts for all TIFAv1.0 4,081 captions (prompts). Most captions have just 1 or no element mismatch, indicating that LLaVA can break down and generate questions for prompts with similar performance to the proprietary GPT-3. \looseness-1
    }
    \label{fig:el_mismatch}
\end{figure}

Across all prompts, TIFAv1.0 has an average of 4.68 elements per prompt and LLaVA-Gen TIFA has 4.13 elements per prompt, indicating an extremely similar number of elements identified for each prompt.
For prompts with mismatched elements, the average number of mismatched elements is only 1.13, and we show counts of prompts with mismatched elements in Figure \ref{fig:el_mismatch}.
The average BLEU \citep{bleu} score between TIFAv1.0 and LLaVA-Gen TIFA questions about the same elements is 0.65, which indicates a high level of similarity between questions covering the same concepts.

\section{Limitations \& Future Work Discussion} \label{app:discussion}

A limitation specific to \method's design is that a single MLLM both generates the evaluation prompts and scores the resulting images. This couples two roles that GQA pipelines keep separate, and it admits a failure mode those pipelines do not have: the evaluator could favor prompts whose satisfaction it is itself well equipped to verify, inflating measured consistency in a way an independent judge would not reproduce. Our results bound this concern without eliminating it. The ground truth in Sections~\ref{sec:mllm_genbench} and~\ref{sec:adaptive} is external to \method---each benchmark's own evaluation pipeline run over its own prompt set---so a scorer that systematically favored its own prompts would have to do so in a way that nonetheless reproduces three independently constructed rankings; and Section~\ref{sec:contribution_1} validates the scorer against human judgment on prompts it did not write. Neither check isolates the effect directly. Doing so would require scoring \method-generated prompts with an independent judge, or collecting human consistency ratings on them, which we leave to future work.

Beyond this, the use of open-source MLLMs as dynamic evaluators for T2I models introduces several limitations to the evaluation process, inherited from various well-known weaknesses of MLLMs. Notably, MLLMs are susceptible to hallucination and implicit bias, generating outputs that appear credible but are not faithfully grounded in the input images \citep{visualhallucinationsmultimodallarge, li2023evaluatingobjecthallucinationlarge}.
This is largely due to their tendency to over-rely on language patterns and statistical correlations inherent in training data, rather than genuine reasoning and understanding of their inputs. This has been systematically observed to undermine MLLM reliability and performance in tasks such as visual question answering and vision-language reasoning, which are crucial to using MLLMs as evaluators \citep{goyal2017making, winoground, eyeswideshut}.
Recent work to mitigate MLLM hallucinations involves fine-tuning models on hallucination-corrected multimodal datasets or contrastive learning to discourage hallucinated text \citep{visualhallucinationsmultimodallarge, hallucinationaugmentedcontrastivelearning}.
Both approaches have yielded significant improvements in hallucination rates for open-source MLLMs, including LLaVA \citep{llava}, one of the models we evaluate in Section~\ref{sec:contribution_1}.
Because \method\ is designed to work with any MLLM, improvements in reducing bias and hallucinations in these models can be directly leveraged. As more robust and reliable MLLMs become available, \method\ can seamlessly incorporate them to deliver more trustworthy T2I evaluations.

Future work could explore leveraging multiple MLLMs as ensembles of T2I evaluators, to investigate whether majority voting among different models leads to more reliable evaluations; inspired by the use of inter-annotator agreement as a consistency measure for human evaluations of T2I models. Furthermore, as advanced reasoning techniques like chain-of-thought become available in multimodal models, incorporating self-reflection and self-verification within the T2I evaluation process could help validate image-text judgments and facilitate the creation of more effective, adaptive evaluation prompts \citep{bagel, zebra_cot}.

\end{document}